\documentclass[12pt]{article}

\usepackage{amsmath, amssymb}
\usepackage{graphicx}
\usepackage{booktabs}
\usepackage{multirow}
\usepackage{natbib}
\usepackage{xurl}
\usepackage{hyperref}
\usepackage{caption}
\usepackage[utf8]{inputenc}
\usepackage{subcaption}
\usepackage{float}
\usepackage{algorithm}
\usepackage{algpseudocode}
\usepackage{geometry}
\usepackage{tabularx}
\usepackage{longtable}
\usepackage{xcolor}

\title{Same Accuracy, Different Gene Panels: How the Hyperparameter Grid Chooses the Reported Explanation}
\author{
Chama Bensmail \\
Biocomputation Research Group\\
University of Hertfordshire \\
\texttt{c.bensmail2@herts.ac.uk}
}

\date{\today}

\begin{document}
\maketitle
\begin{abstract}
Which biological findings survive into a reported explanation depends on how the population of models was constructed; a choice that predictive validation alone does not reveal. The dependence is systematic, measurable, and large enough to change what a reported gene panel contains.

We fitted $10^4$ L2-regularised logistic-regression models to a TCGA pan-cancer tumour-versus-normal task: 100 regularisation strengths on each of 100 train--test splits, with the patients fixed within each split. Strongly and weakly regularised models differ in mean accuracy by only 0.21 percentage points, yet rank different genes highest, and the change follows the regularisation strength. We then simulated the common practice of averaging explanations across a hyperparameter grid and reporting the top 50 genes. The reported panel shares 41 of its 50 genes with a single fitted model---but which model is set by where the practitioner centred their grid (Spearman $0.92$, 350 simulated practitioners), and the remainder is composed by the averaging: for 43 of 50 practitioners the panel combines the genes of strongly and weakly regularised models more fully than any model they fitted, so the reported explanation belongs to no model whose accuracy was established. Genes favoured only by the most strongly regularised models are largely lost: a conventional grid recovers a quarter of them, and HMGCS2, among the top 50 genes of strongly regularised models in 73 of 100 splits and of weakly regularised models in none, appears in none of 50 simulated reports.

Reconstructing a published biomarker workflow and refitting it on 20 resamples of its cohort shows a different failure: no refit comes close to the averaged panel---the closest contains 23 of its 50 genes, and no gene in it appears in all 20 refits. We define a panel's \emph{model-relative provenance}---how much of it any single fitted model contains, as distinct from the workflow provenance that records how an analysis was run---and give the measurements that make the model population, the choice that selects from it, and that provenance reportable.
\end{abstract}

\section{Introduction}

Which biological findings survive into a reported explanation depends on how the
population of models was constructed. That dependence is the subject of this
paper. It is not a claim that explanations are unstable, which is established;
it is a claim that the instability has a coordinate, that a practitioner sets
that coordinate when they choose a hyperparameter grid, that predictive
validation under a fixed accuracy tolerance does not determine it, and that
reporting the grid alone does not reveal its consequences. We measure the dependence, identify the coordinate for regularised
linear models, and show what it costs in reported biological content.

Machine learning models are increasingly deployed in domains where the
explanation, not merely the prediction, is the deliverable. In high-dimensional
genomics, for instance, a transcriptomic classifier may reach near-identical
predictive accuracy under many admissible training configurations while
assigning importance to different groups of genes; the resulting
feature-importance vector is then read as a biological finding and frequently
transformed into a pathway-level narrative. Despite the fact that practitioners
routinely retrain models during development and performance tuning,
interpretability is typically assessed using a single trained instance---or by
averaging explanations across runs---rather than by explicitly examining whether
independently selected, equally accurate models in fact agree. We take
this aggregate as our object of study, and ask what it conceals: whether the
models being summarised share one explanation, or whether the consensus averages
over several equally accurate but mutually distinct ones. Reporting one
aggregate as the explanation implicitly treats the former as the case, as though
high predictive accuracy corresponded to a single, well-defined way of reasoning \citep{lipton2018myth,rudin2019stop}. That
assumption is rarely tested.

It has been challenged from several directions: the Rashomon
effect, under which distinct models reach similar accuracy on different internal
logics \citep{breiman2001statistical,marx2020predictive}; underspecification,
under which a pipeline admits many equally accurate solutions and selects among
them arbitrarily \citep{d2022underspecification}; the instability of attributions
under retraining, feature correlation and perturbation
\citep{kindermans2017reliability,nogueira2018stability,kumar2020problems,frye2020missingness};
and a line of work treating variable importance as ranging over the set of
well-performing models rather than as a property of one
\citep{fisher2019all,donnelly2023rashomon}. Collectively these establish that
explanation stability cannot be assumed and that importance is better reported as
a range than a point. Section~\ref{sec:related} sets out what they do and do not
settle.

Three things they do not establish are what this paper supplies. First, that work
estimates importance over the admissible set and, in places, describes clustered
importance patterns within it \citep{dong2020exploring}; it does not ask how a
reported aggregate depends on where a practitioner samples that set. We show that
for regularised linear models the set has a \emph{coordinate},
that explanatory content varies systematically along it, and that which member a
practitioner reports is predicted by their grid centre, with rank correlation
$0.92$ across 350 simulated practitioners. Second, that literature stops at the
model population; practice does
not, and the aggregation step that follows retraining is where reported content
is actually lost---we show that it resolves none of the disagreement, that what
it retains depends on how the grid was placed, and that what it reports is in
part a combination no fitted model contains. Third, the endpoint
here is biological: not how much importance varies, but which genes reach the
reported panel and which do not.

We therefore treat the explanations a pipeline produces as a population rather
than a single object, ask whether that population is homogeneous, and then ask
what the conventional summary of it actually reports. Our analysis examines
which explanatory content survives aggregation, which training choices determine
that survival, and how much of the reported panel is supported by any single
fitted model. These measurements expose choices and losses that predictive
validation alone does not reveal, making them available for explicit reporting
and audit
\citep{arrieta2020explainable,selbst2018intuitive,ecegai2023aiact}.

We assess the provenance of a reported panel by measuring the largest fraction of its features contained together in any single constituent model’s attribution panel. This distinguishes support for individual features across a population from their joint occurrence within one fitted explanation. The measure requires neither clustering nor a partition and can be applied across model classes when explanations are expressed as comparable feature sets. This is provenance relative to the fitted model population, not provenance in the sense of workflow or metadata lineage, which records the software versions, parameters and annotation releases behind a result \citep{abassi2026emma}. A panel can be fully reproducible in that second sense and still be contained in no fitted model. Combined with the identity and pipeline setting of the best-matching fit, it reveals how closely the reported panel corresponds to a constituent explanation and how that correspondence depends on the practitioner’s training choices.

Applying this to transcriptomic classification on public cancer-genomics data,
we report four findings. First, on a TCGA pan-cancer tumour-versus-normal task,
L2-regularised logistic-regression classifiers across a regularisation path are predictively
interchangeable within a declared one-percentage-point tolerance---mean held-out
accuracy $0.9838 \pm 0.0027$---yet their attributions vary systematically with the
regularisation parameter: explanatory content changes gradually over two to
three decades of $C$, between a strongly regularised (low-$C$) and a weakly
regularised (high-$C$) end whose mean accuracies differ by $0.21$ percentage
points, and where a fit sits on that path follows $C$, with patient membership
held fixed within each split. Second, the two ends report different genes, reproducibly:
\textit{HMGCS2} is among the top 50 genes of the low-$C$ regime in 73 of 100
splits and of the high-$C$ regime in none, and \textit{CCK} in 2 and 87. The differences are reproducible at the level of individual genes, not of enriched pathways.

Third, simulating the conventional workflow in which a practitioner sweeps the
regularisation grid and averages the resulting explanations, the reported panel
shares 41 of its 50 genes with its closest fitted model (42 under SHAP), but which model is
closest is set by where the grid was centred rather than by any deliberate choice: in logarithmic units the
correlation between grid centre and best-matching model is $0.88$ (Spearman
$0.92$) across 350 individual practitioners. \textit{HMGCS2} appears in none of
50 simulated reports from a conventional grid, which recovers on average about 3 of the 11 genes found only in the most strongly regularised models ($C \le 0.016$); centring the grid on those models raises this to about 9 of 11. What the panel does not share with its
closest model is composed by the averaging: for 43 of 50 practitioners the panel
holds more of both regimes' content at once than any of the models they
averaged. Replacing the average with selection by recurrence across fits does not avoid the loss in the conditions tested. A seed-varied deep network on the same data shows a third
structure---most fits agree on one direction and a minority invert it in their
class-1-logit attributions---and a
seed-varied gradient-boosted ensemble concentrates on one explanation, which
recovers most of the genes specific to the low-$C$ regime of the regularisation
path; together these bound when the phenomenon occurs and show that the
low-$C$ content is not specific to one model class.

Fourth, reconstructing a published biomarker workflow and refitting it across
20 resamples of its cohort yields models whose explanations are
dispersed: of the 50 genes a standard workflow would report,
the most similar refit contains 23 and none appears in all 20. Holding the
evaluated patients fixed does not restore agreement, so the disagreement is
between the models rather than between the patients they are evaluated on. Where
the regularisation path yields a panel largely contained in a model the
practitioner did not choose, this population yields one that no refit contains
beyond 23 of its 50 genes; neither case is visible in the published artefact
without a population-level check.

\section{Related Work}
\label{sec:related}
\paragraph{Multiple accurate models.} The Rashomon effect observes that
different models can achieve similar predictive performance while relying on
different internal reasoning
\citep{breiman2001statistical,semengue2019rashomon}. Predictive multiplicity has
since been studied in settings such as fairness and unconstrained risk
minimisation \citep{umbach2023predictive,marx2020predictive}, and it is not a
reliable indicator of explanatory multiplicity: models of near-identical
aggregate performance can give conflicting explanations of the same decision
\citep{brunet2022implications}; and work on
underspecification shows that modern training pipelines may admit many solutions
that perform equally well \citep{d2022underspecification}. A related line incorporates this multiplicity into post-hoc
explanation by operating over Rashomon sets: \cite{cavus2025rashomonpdp}
aggregate partial dependence profiles across near-optimal models and report
uncertainty bands to highlight disagreement. Such approaches acknowledge that
multiple admissible models exist, but represent the resulting multiplicity as
uncertainty around a single aggregated explanation. We study multiplicity within
one model class and one admissible pipeline, and ask whether that aggregation is
warranted before it is performed.

\paragraph{Importance across a model set.} Closest to this work is a line that
treats variable importance not as a property of one fitted model but as a
quantity ranging over the set of well-performing models. Model class reliance
bounds a variable's importance across that set \citep{fisher2019all}, and
Rashomon importance distributions estimate its distribution across the set and
across data perturbations, yielding variable-importance estimates that are
stable where single-model estimates are not \citep{donnelly2023rashomon}.
Variable importance clouds go further, mapping the joint importance structure of
the good-model set and, in their examples, clustering it into groups with
representative models \citep{dong2020exploring}.
\citet{laberge2023partial} take the opposite route to summarisation: rather than
average attributions, they retain only the importance statements on which every
model within a loss tolerance agrees, and report the resulting partial order;
much of what a single averaged ranking asserts is not supported by that
consensus. These
establish that importance is better reported as a range than a point, and supply
estimators for that range. We ask two different questions. First, given that the
set can have internal structure, we ask whether that structure has a coordinate
a practitioner controls: for regularised linear models it does, and content
varies systematically along it, so that the admissible range is an ordered path
whose position a practitioner fixes when they choose a grid. Second, these
methods characterise the model set, whereas what a genomics paper publishes is a
summary produced by averaging or recurrence selection over whichever subset of
that set the practitioner happened to fit. We measure what that step does.
Panel provenance therefore differs from consensus over the admissible set: it
asks not which statements every admissible model supports, but whether the panel
a practitioner actually reports is contained in any single fitted model, and
which one---a containment test against fitted models, tied to where the grid
sampled the path.

\paragraph{Unstable explanations.} Explanation methods such as SHAP, LIME and
gradient-based attribution exhibit variability under retraining, dataset
perturbation, or feature correlation
\citep{kindermans2017reliability,nogueira2018stability,kumar2020problems,agarwal2021uncertainty}.
These works concern instability in individual attribution values or feature
rankings; in transcriptomic prediction specifically, optimiser and
regularisation choices alter selected gene signatures without commensurate
changes in predictive performance \citep{crawford2024optimizer}. On TCGA
directly, classifiers of different families reach similar accuracy from
substantially different top-ranked genes, and comparable performance remains
achievable with lower-ranked ones \citep{bontonou2025comparative}. That equally
accurate transcriptomic models disagree on genes is therefore established; what
this paper adds is the coordinate along which they disagree, and what the
reported summary does with that disagreement. A further line
treats explanation non-uniqueness as an object of study rather than incidental
noise---as run-to-run variance in SHAP explanations
\citep{hwang2026multiplicity}, as non-identifiable circuit recovery in
mechanistic interpretability \citep{meloux2025everywhere}, or as mechanisms that
must be treated as estimators with variance
\citep{mi2025variance,modularjets2025}.
\citet{hwang2026multiplicity} separate model-related from explainer-intrinsic
sources of SHAP variance, but fix hyperparameters after tuning; how the grid
itself, or the aggregation that follows it, determines the reported explanation
is not examined. We ask a prior question: before
summarising explanation variability as variance around a single explanation, one
should check whether the population is concentrated around one.
\paragraph{Auditing model claims beyond predictive success.} A parallel line
of work asks whether a model's apparent reasoning is warranted by its predictive
success, and finds that it often is not. CellAudit separates whether an input
can enter a model's computation, whether its predictions depend on that input,
and whether the dependence improves prediction; it reports a perturbation-response
predictor whose outputs were invariant to replacing the compound it was designed
to use, and whose accuracy matched a predictor that ignored the compound
\citep{li2026discover}. In matrix factorisation, high reconstruction quality and
agreement across restarts do not guarantee recovery of the generating factors
\citep{ma2026reconstruction}. In reinforcement learning, rule sets extracted
from independently trained policies can overlap substantially while the policies
act differently \citep{liu2026auditability}. These studies audit single models,
or agreement between fits. None asks whether a reported aggregate over a model
population is supported by any member of that population, which is the property
measured here.

\paragraph{Aggregation as standard practice in omics.} In high-dimensional
genomics the instability of single-instance explanations is widely recognised,
and a substantial body of practice has grown up specifically to manage it:
rather than report one model's attribution, practitioners aggregate explanations
or selected genes across many models and report the consensus as the biological
finding. This aggregation takes several standard forms. Ensemble feature
selection combines feature rankings across bootstrap resamples or
cross-validation folds into a single robust signature
\citep{saeys2008robust,abeel2010robust}, an approach that consensus nested
cross-validation formalises by retaining features that recur across inner folds
\citep{parvandeh2020consensus}. More recent explainable-AI workflows average
absolute SHAP values across cross-validation folds to obtain a ``robust''
feature-importance estimate \citep{unitednet2023,multideconv2025}, rank genes by
mean attribution across folds or models \citep{roy2024integrative}, or select
final biomarkers by frequency of selection across a hyperparameter grid or
across multiple algorithms \citep{frontiersdehp2026,firat2026uveitis}. A
parallel convention projects the aggregated gene-level attribution onto curated
gene sets, reporting a pathway-level enrichment narrative in place of the
unstable gene list \citep{zhao2026deepdtf}.

Across these variants the aggregate is usually read as a stable summary of
the models behind it. Some consensus procedures are designed to estimate
something narrower---stability selection, for instance, targets selection
probabilities with error control \citep{meinshausen2010stability}---and none
requires every model to implement one explanation. The risk we examine is
interpretive: reading an aggregate as representative of its constituent models,
or as an explanation some fitted model supports, without checking either
property. We test two such procedures on a controlled population: signed
averaging of attribution vectors (Section~\ref{sec:result3}) and
frequency-thresholded top-$N$ selection, the recurrence rule used in stability
selection (Section~\ref{sec:stability}).

That reading can fail in two ways, and they license different conclusions. If the population carries different content at different settings of a pipeline parameter, an aggregate may represent one of those settings rather than the population, with which one decided by how the sweep was placed. If the population shows little directional agreement, an aggregate may correspond to no fitted model at all. We test the first on a regularisation path (Section~\ref{sec:result3}) and the second on a reconstructed published workflow (Section~\ref{sec:lewis}).

\paragraph{The gap.} What this literature does not establish is what the
aggregation step that follows retraining does with the variation it
documents---whether it preserves the disagreement, resolves it, or discards it
while leaving a summary that appears well-supported. This matters beyond
interpretability research. Structured
assurance methodologies such as AMLAS and the ``BIG Argument'' for AI safety
build an explicit chain of justification from system behaviour back to the
trained model \citep{habli2025big,hawkins2021amlas}, and are therefore sensitive to any step that breaks traceability. Either failure
would break it, and each in a different way. A reported panel that no fitted
model largely contains weakens the link between explanation and model. A
panel largely corresponding to one model selected by an undocumented sweep parameter is
in one respect harder to audit, because the link is largely intact and verifiable and
nevertheless points at an arbitrarily chosen member of a population whose other
members were equally admissible, while the remainder of the panel traces to the
averaging step rather than to any model. An assurance argument that traces a claim back
to that model will largely succeed, and is unsound if it also claims the explanation is
robust or representative of the admissible models, which the trace does not
establish. Panel provenance is therefore one traceability property among
several, not a certificate of validity. As with auditability of learned
policies, where separately testable properties do not license one another
\citep{liu2026auditability}, a panel contained in a fitted model can still
misrepresent the admissible population (Section~\ref{sec:result3}).
\section{Problem Formulation}

Formally, consider a model class $\mathcal{A}$ trained on a fixed dataset
$\mathcal{D}$ using a fixed admissible training pipeline $\mathcal{P}$, which
includes preprocessing, model architecture, optimisation procedure, and
admissible hyperparameter configurations. Standard interpretability workflows
examine a single trained instance $\theta \sim \mathcal{P}(\mathcal{A},
\mathcal{D})$ and its corresponding attribution vector $\Phi(\theta)$. In
contrast, this work asks: when $\mathcal{A}$ is instantiated repeatedly under the
same admissible pipeline---via admissible hyperparameter and model-selection
choices, with random seeds and resampling as additional perturbations---do the
resulting attribution vectors describe one explanation, or several?

Let $\{\theta_r\}_{r=1}^{R}$ denote $R$ model instances produced by pipeline
$\mathcal{P}$ on $\mathcal{D}$, and let $\mathbf{e}_r = \Phi(\theta_r) \in
\mathbb{R}^d$ denote the corresponding explanation vectors, where $d$ is the
number of attributed features. The set $\{\mathbf{e}_r\}_{r=1}^{R}$ defines an
empirical distribution over explanation space induced by admissible training and
model selection. We refer to this as the \emph{explanation population} of the
pipeline.

We characterise this population along three axes:

\begin{enumerate}
\item whether the population is homogeneous, separates into distinct
      explanatory regimes, or varies continuously with an admissible pipeline
      parameter;
\item how much empirical support each regime receives, and which pipeline
      choices determine which regime a model instance occupies;
\item what a conventional aggregate over the population reports, relative to
      what the constituent models individually contain---and whether any
      constituent model supports the aggregate at all.
\end{enumerate}

We use the term \emph{regime} rather than cluster or basin deliberately. In the
pipelines studied here the population is not always a set of separated point
clouds. On the regularisation path it is better described as a continuous
trajectory through explanation space with coherent regions at either end and a
narrow transition between them. What matters for practice is not the geometry of
that trajectory but whether the regions carry different content and whether a
single summary can represent them. We therefore avoid claims about separation
that the data do not support, and identify regimes by the pipeline parameter
that determines membership rather than by the geometry of the point cloud alone.
Where the parameter that determines membership is continuous, we treat position
on its range, rather than membership of a resolved region, as the primitive.
This costs little---the content varies with the parameter either way---and it
reduces dependence on a partition procedure whose granularity is a property
of the attribution method rather than of the pipeline
(Section~\ref{method:positions}).
In practice, we repeat this analysis across multiple train--test splits of
$\mathcal{D}$ to assess whether explanatory structure is reproducible under data
resampling.

\section{Methods}

We call this population-level analysis EvoXplain. It characterises the
explanations produced by a declared training pipeline and measures what their
reported summary retains and how closely it corresponds to individual fitted
models.

The workflow is
simple: repeatedly instantiate the admissible pipeline, extract explanations,
represent them in a common space, characterise the resulting population, and
compare what a conventional summary of that population reports against what the
individual models contain.

Four terms recur below and are kept distinct. A \emph{region} is any group of
fits that a lens's partition resolves (Section~\ref{method:regimes}). A
\emph{regime} is a region whose membership is determined by a pipeline parameter;
on the regularisation path these are the two regions LIME resolves, the strongly
regularised (low-$C$) and the weakly regularised (high-$C$). A \emph{position} is
a range of the parameter cut at fixed thresholds, independent of any partition
(Section~\ref{method:positions}). A \emph{profile} is the mean unit-normalised
attribution vector of the fits in a region or position.

\subsection{Repeated Training and Attribution Sampling}
\label{method:setup}

Let $\mathcal{A}$ denote a model class (for example, regularised logistic
regression or a gradient-boosted ensemble). For each run $r = 1,\dots,R$, we
train $\mathcal{A}$ under an independently instantiated admissible training
pipeline, while keeping the data split and preprocessing fixed. A pipeline
instantiation may involve stochastic sources (optimiser initialisation, row and
column subsampling) and/or admissible hyperparameter configurations.

After each training run we compute feature-level attributions using a chosen
interpretability method, yielding an attribution vector $\mathbf{e}_r \in
\mathbb{R}^d$, where each component is the attribution assigned to one input
feature. EvoXplain collects these vectors,
\[
E = \{\mathbf{e}_r\}_{r=1}^{R},
\]
and treats them as samples from the explanation population induced by repeated
admissible instantiation of the pipeline. EvoXplain does \textbf{not} aggregate
predictions or construct ensembles; it analyses explanatory structure arising
solely from admissible variation within a single model class and data split.

The results reported here use SHAP and LIME as the two attribution lenses,
selected because they are the most widely deployed model-agnostic attribution
methods in applied settings.

Reporting two lenses allows us to separate properties of the attribution method
from properties of the underlying pipeline. To ensure comparability across runs,
all attribution vectors are computed using the same interpretability method,
background reference, and preprocessing configuration.

\paragraph{Which patients are explained.} Within each split, every fit is
explained on the same 200 held-out patients, chosen near the decision boundary
of a reference random forest (predicted probability between 0.45 and 0.55). A
fit's explanation is the average of its attributions over those patients.

\paragraph{What each lens explains.} SHAP \citep{lundberg2017shap} explains the
model's score before it is turned into a probability (for logistic regression,
the log-odds); LIME \citep{ribeiro2016lime} explains the predicted probability
itself. For logistic regression and the boosted trees the two rise and fall
together, so the distinction does not matter; for the neural network they can
disagree in sign. We therefore never compare attribution values across lenses:
every comparison is between fits explained by the same lens, and the two lenses
are compared only on the structure each finds. Exact explainer settings are
given in Appendix~\ref{sec:implementation-appendix}.

\paragraph{Randomness in the explainer.} SHAP for logistic regression is
deterministic: a given model always receives the same explanation, so variation
between SHAP explanations on the regularisation path comes from the models.
LIME samples random perturbations and carries some variance of its own
\citep{hwang2026multiplicity}.

\paragraph{Where variation between fits comes from.} L2-regularised logistic
regression has a single optimum for given data and $C$, so on the
regularisation path fits differ only because $C$ differs. Boosted trees are
deterministic unless rows and columns are subsampled; we enable subsampling, so
different seeds give genuinely different models, and we check that they do.

\paragraph{Preprocessing.} The 1{,}000-gene variance filter and per-gene scaling
are computed once on the full cohort, before any split. They use no labels, so
no outcome information reaches the test set, but test patients do contribute to
the scaling. All comparisons are conditional on this choice; train-only
preprocessing, which could shift the path itself, was not tested.

\subsection{Explanation Space Representation}

Each attribution vector $\mathbf{e}_r \in \mathbb{R}^d$ is a point in a
$d$-dimensional explanation space $\mathcal{E} \subset \mathbb{R}^d$, where $d$
is the number of input features. Because all attribution vectors are computed
over the same feature set and under the same preprocessing and attribution
configuration, they are inherently aligned and directly comparable across runs.

EvoXplain analyses explanation vectors in their native attribution space,
preserving the feature-wise structure produced by the interpretability method.

\subsection{Distance Metric}

To compare explanation vectors, EvoXplain uses cosine distance,
\[
d(\mathbf{e}_r, \mathbf{e}_s)
=
1 -
\frac{\mathbf{e}_r \cdot \mathbf{e}_s}
{\|\mathbf{e}_r\|_2 \, \|\mathbf{e}_s\|_2}.
\]
Cosine distance measures differences in attribution direction rather than
magnitude, ensuring that comparisons reflect differences in explanatory logic
rather than global scaling of attribution values. In practice all analysis is
performed on unit-normalised explanation vectors, making cosine distance
equivalent to half the squared Euclidean distance in the transformed space.

\subsection{Characterising the Explanation Population}
\label{method:regimes}

We characterise the population along two axes: whether it separates into
distinct regions, and whether any such region is internally coherent. These are
different questions and a single statistic answers neither.

\paragraph{Partition.} To assess whether explanations separate, we evaluate
candidate partitions using $k$-means for $k \in \{2,\dots,K_{\max}\}$, compute
the silhouette score $S(k)$, and select
\[
k^\star =
\begin{cases}
\arg\max_{k} S(k) & \text{if } \max_k S(k) \ge \tau \\
1 & \text{otherwise.}
\end{cases}
\]
Two further conditions are applied before a candidate $k$ is scored, and we
state them because both can return $k^\star = 1$ on a population that is not
homogeneous. A candidate $k$ is discarded if any cluster holds fewer than two
members, and if the cosine between any pair of cluster centroids exceeds
$0.99$, the latter to prevent $k$-means from reporting a split that separates
magnitude rather than direction. A population of fewer than $K_{\max}+1$
members, or with variance below $10^{-9}$, returns $k^\star = 1$ without a
search. In this work $K_{\max} = 8$ and $\tau = 0.15$, so $k^\star = 1$ means
``no partition survived those conditions'', not ``the population is a single
coherent region''; the coherence measure below is what answers the latter.
We report the full silhouette profile $S(k)$ rather than $k^\star$ alone. The
distinction matters: a profile that increases monotonically to the largest $k$
tested, or stays nearly flat across a range of $k$, indicates that $k$-means is
segmenting a connected structure and that $k^\star$ is set by the search range
or by noise, whereas a clear interior maximum indicates a resolved partition.
$k^\star$ is in general lens-dependent, since different
attribution methods resolve the same population at different granularity, so we
do not treat any single partition as canonical. Where a partition is reported,
its status is established by the profile, by the determinism check below, and
by the content of the resulting regions. The empirical support of region $j$ is the fraction of runs
assigned to it, $w_j = |C_j| / R$ with $C_j = \{r : c_r = j\}$.

\paragraph{Coherence.} The count $k^\star$ records how many regions of
concentration a partition identifies, but not whether any such region is itself
coherent: a $k^\star = 1$ outcome conflates a single \emph{concentrated}
population with a \emph{dispersed} cloud possessing no preferred direction,
as neither need yield an accepted partition. To separate these cases we report, for
each region $j$, a directional coherence
\[
\rho_j \;=\; \left\lVert \frac{1}{|C_j|} \sum_{r \in C_j}
\frac{\mathbf{e}_r}{\lVert \mathbf{e}_r \rVert} \right\rVert \;\in\; [0,1],
\]
the mean resultant length of the unit attribution vectors in the region. A
concentrated region has $\rho_j \to 1$: member attributions agree in direction
and sign. A structureless cloud has $\rho_j \approx 0$, with directions
showing little directional agreement, so that no single direction is
reliably recovered; a small $\rho_j$ is not by itself proof of isotropy, since
symmetric structured populations can also produce it.

We note what $\rho_j$ does \emph{not} add. For $R$ unit vectors it is a
monotone rescaling of the mean pairwise cosine $\bar{c}$,
\[
\rho^2 \;=\; \frac{1 + (R-1)\,\bar{c}}{R},
\]
so the two carry identical information about overall coherence and neither can
resolve structure the other cannot. We report $\rho_j$ because it is bounded,
interpretable as a length, and comparable across regions of different size, not
because it is more sensitive. Where a population is concentrated overall but
contains a sign-reversed minority, both summaries are blind to the distinction,
and it is the \emph{distribution} of pairwise similarities---its minimum, the
presence of negative pairs, and the coherence of the subgroups---that separates
the two cases. We use that evidence wherever the distinction matters.

\paragraph{Determinism check.} Where a population separates, we ask what
determines membership. Within a single train--test split every model instance
sees the identical patient sample, so any variable that is constant within a
split cannot explain differences between instances. If region membership is a
deterministic function of one pipeline parameter---on the regularisation path, a
single threshold in the regularisation strength $C$---then the separation is a
property of the pipeline, and sample composition, class balance and cohort
heterogeneity are excluded by construction rather than by statistical
adjustment. We report the fraction of splits in which membership is monotone in
the varied parameter, and the location of the transition. 
To locate where content changes along a continuous parameter, we place each fit
on the axis joining the mean unit explanation vectors $\mathbf{a}$ and
$\mathbf{b}$ of the ten fits at either end of its split's path,
$s_r = (\mathbf{e}_r - \mathbf{a})\cdot(\mathbf{b}-\mathbf{a})/\lVert\mathbf{b}-\mathbf{a}\rVert^2$,
which runs from $0$ at the low end to $1$ at the high end; attribution noise
scatters $s_r$ but does not bias it. We report the parameter value at which the
running-median progress reaches $0.5$, and the width between $0.1$ and $0.9$ in
decades.
\subsection{Position on a Parameter Path}
\label{method:positions}
 
The partition above answers how many regions a lens resolves, and that count is
lens-dependent by construction (Section~\ref{sec:result2}). Where the pipeline
parameter being varied is continuous, a second decomposition is available that
does not depend on any partition procedure, and the aggregation analyses use it
in preference.
 
We divide the parameter's range at fixed thresholds into \emph{positions}, and
characterise each position exactly as a region is characterised: its profile is
the mean unit-normalised attribution vector of the fits falling in it, and its
\emph{reproducible core} is the set of features entering that profile's top $N$
in at least a fraction $\theta$ of the data splits. A feature is \emph{exclusive}
to a position when it belongs to that position's core and to no other's. Core
size measures how much of a position's content recurs at all; exclusive core
size measures how much of it distinguishes that position from the rest of the
path. We use $\theta = 0.7$ and $N = 50$ throughout.
 
Two properties of this decomposition matter for what follows. It introduces no
clustering of its own, so a result computed on it depends on the lens-specific
partitions only through the boundary positions, which are tested against
alternatives below; and it is defined on the parameter rather than on the explanation
vectors, so positions remain comparable across attribution methods. The
thresholds used for the regularisation path ($C \approx 0.016$ and
$C \approx 0.081$) are taken from Sections~\ref{sec:result1}
and~\ref{sec:result2} and are therefore estimates rather than free choices;
Section~\ref{sec:alternatives} reports the analyses repeated with thresholds
chosen without reference to any partition.
 
\subsection{Explanation Status}

Combining the three diagnostics yields four qualitatively distinct outcomes for
an explanation population, which we report as its \emph{status}:

\begin{itemize}
\item \textbf{Single}: $k^\star = 1$ with high $\rho$. The pipeline yields one
      concentrated explanation.
\item \textbf{Plural}: $k^\star > 1$ with each region coherent and its membership
      attributable to an identifiable pipeline choice.
\item \textbf{Dispersed}: $k^\star = 1$ with low $\rho$. No preferred direction
      exists; any reported panel, including an average, corresponds to no
      concentrated explanation.
\item \textbf{Indeterminate}: the diagnostics disagree across lenses or the
      population is too small to resolve. The correct action is to abstain from
      a singular explanatory claim.
\end{itemize}

Panel provenance is measured separately (Section~\ref{method:provenance}).

\subsection{Region Profiles and Their Content}

For each region we form a profile as the mean unit-normalised attribution vector
of its members, and rank features by absolute profile weight. Region content is
reported at gene level: which genes enter each profile's top 50, in how many
splits, and how far each position's reproducible core is exclusive to it
(Table~\ref{tab:cores}). Gene-level content is measured directly on ranked
attributions and involves no ontology.

\paragraph{Enrichment as a test, not a description.} A practitioner would
typically carry such profiles into over-representation analysis. We run that
analysis to test whether gene-level differences assemble into distinct
pathways, not to describe the regions. Each profile's top 50 genes are tested
against Gene Ontology, KEGG and Reactome, per split and per region, with no
averaging of profiles across splits, and term--gene intersections are retained,
since several terms in these data are driven by the same few genes. The
background is declared for each analysis: the 1{,}000-gene modelling feature
space for the regularisation-path analyses, since every profile is drawn from
it, and the assay's 22{,}835 measured genes for the retrospective audit
(Section~\ref{sec:lewis}). Because the feature space is itself enriched against
the genome-wide domain, we measure the power of the feature-space test by
planting gene sets of known size into regime panels, and complement it with a competitive gene-set test that uses the whole
attribution profile rather than a top-50 cut. In each split, the difference in
mean absolute attribution between the two LIME regimes is $z$-scored across the
feature space; each GO, KEGG or Reactome set of 10--200 feature-space genes
(746 sets) is scored by the mean $z$ of its members against random sets of the
same size, a PAGE-style test \citep{kim2005page} with finite-population
variance, and sets are called at Benjamini--Hochberg $q < 0.05$ within split.
Because the regimes are defined by clustering these vectors, that they differ is
true by construction; the test asks only where the difference concentrates. Exact
settings, database release and request-completeness checks are given in
Section~\ref{sec:implementation-appendix}.

\paragraph{Independent corroboration.} For the genes that distinguish regions we
record whether the direction of effect is supported by independent evidence
outside the cohort, prioritising immunohistochemical, proteomic and functional
studies over transcriptomic reanalyses of the same data, and compare it with the
within-site direction in this cohort (Table~\ref{tab:tissue}). Genes without
such support are reported as uncorroborated. Corroboration establishes a gene as
a tumour--normal marker; it does not establish that the region recruiting it is
biologically preferable, and we describe recurrent genes as reproducible model
interpretations throughout.

\subsection{Simulating the Reported Explanation}
\label{method:phantom}

The preceding diagnostics describe the population. To assess what current
practice reports about it, we simulate the conventional workflow directly.

Each simulated practitioner draws a subset of the available data splits, scans a
hyperparameter grid within each, and averages the resulting unit-normalised
explanation vectors into a single reported vector. The average is taken over
the \emph{signed} vectors and the result is ranked by magnitude, so a feature
whose attributed sign varies across the fits averaged can cancel; this is one
common rule. Workflows that average absolute attributions instead, as in
mean-$|\mathrm{SHAP}|$ global importance, are not reproduced here: taking
absolute values first would suppress any sign cancellation, so the results of
Section~\ref{sec:result3} are specific to signed averaging. The top $N$ features of that vector form their
reported panel. No models are refitted: every vector entering the
average is drawn from the already-computed population, so the simulation isolates
the effect of the aggregation step rather than introducing new variation.

A practitioner's constituent set is the collection of fits their sweep visits:
for the regularisation path, the fit nearest each grid point in $\log C$ in each
drawn split, so that every simulated grid is snapped to the 100-point fitted
path and a grid extending beyond $[10^{-3}, 10^{3}]$ is clipped to it;
for a resampling population, the drawn refits themselves. We write $P$ for the
reported panel and $T_1,\dots,T_M$ for the top-$N$ feature sets of the $M$
constituent fits. Every measurement below is computed on this pair, so that
results from different populations and different aggregation rules are directly
comparable. The simulation is repeated 50 times per condition with independently
drawn splits; reported figures are means over those repetitions.

\subsection{What the Report Retains, and What Supports It}
\label{method:provenance}
 
Two distinct properties of a reported panel are measured.
 
\paragraph{Content.} For position $j$ with exclusive core $X_j$, the content the
panel carries is $|P \cap X_j| / |X_j|$. Because exclusive cores can be small,
we report $|X_j|$ alongside every such figure and omit positions whose exclusive
core is too small to support a proportion.
 
\paragraph{Panel provenance.} The model-relative provenance of a panel, which
we call panel provenance, is the fraction of it contained in the single
best-matching fit that produced it,
\[
\mathrm{prov}(P) \;=\; \max_{i \le M} \frac{|P \cap T_i|}{|P|},
\]
so that $\mathrm{prov}(P) = 1$ means some fitted model contains the entire
reported set. We interpret it against the baseline agreement between two
constituent fits, $\mathbb{E}_{i \neq j}\,|T_i \cap T_j| / N$, and where the
constituent set is indexed by a pipeline parameter we also record the parameter
value of the maximising fit.
 
The absolute value of $\mathrm{prov}(P)$, not its ratio to that baseline, is the
quantity we report, and the reason is a property of dispersed populations rather
than a convention. A high ratio need not imply strong absolute support: in Section~\ref{sec:lewis} it is
$2.6$ against $1.05$ (LIME; $0.91$ under SHAP) for the far more coherent population of
Section~\ref{sec:result1}. We report both figures wherever the comparison is
made, for exactly that reason.

\paragraph{Joint coverage.} Where the population carries more than one region
signature, we also record a panel's joint coverage: the smaller of the
fractions of each signature it contains. We compare it with the largest joint
coverage attained by any of the practitioner's own constituent fits. A panel
whose joint coverage exceeds that of every constituent reports a combination of
content that no model the practitioner fitted contains; every gene in it may
still be supported by some fit.
 
\paragraph{Sensitivity to the sweep.} Where the pipeline parameter is
continuous, we additionally vary the practitioner's grid and repeat the
simulation. Two families are used. A \emph{span} sweep fixes the number of grid
points and varies the interval endpoints. A \emph{centre} sweep fixes the grid's
width in decades and slides its centre along the path; for each centre we record
the parameter value of the best-matching fit, and report the correlation between
that value and the grid centre in logarithmic units. A high correlation means
the reported panel is determined by where the practitioner placed their sweep.
\subsection{Aggregation Rules Compared}
\label{method:rules}
 
Averaging unit-normalised vectors is one aggregation rule among several in use,
and both the scaling applied before averaging and the alternative of selecting
by recurrence are varied explicitly.
 
\paragraph{Scaling.} Four rules are compared, differing only in how each vector
is scaled before the mean is taken: division by the $\ell_2$ norm (the default
throughout), by the $\ell_1$ norm, by the largest absolute component, and a rank
aggregation in which each fit contributes the within-fit rank of every feature
and the panel is the features of smallest mean rank. The first three are
per-vector rescalings and the fourth is invariant to any of them, so each fit's
own top-$N$ set is identical under all four and only the aggregate differs. To
characterise how concentrated a fit's attribution is, we report the
participation ratio $\big(\sum_i |v_i|\big)^2 / \sum_i v_i^2$, the effective
number of features carrying the explanation.
 
\paragraph{Selection by recurrence.} In place of averaging, a panel may be
assembled from the features appearing in the top $N$ of at least a fraction
$\pi$ of the constituent fits. We report $\pi \in \{0.3, 0.5, 0.7, 0.9\}$,
spanning the thresholds used in practice, together with the mean per-feature
support of the resulting panel---the mean fraction of constituent fits
containing each of its members.
 
\paragraph{Matched-length control.} Panels selected at a high $\pi$ are short,
and a shorter panel is easier for a single fit to contain, so provenance can be
increased by length alone. Every recurrence-selected panel is therefore compared
against an averaging panel truncated to the same number of features. Comparisons
between rules are made only at matched length.
 \subsection{Separating Model Variation from Sample Variation}
\label{method:patientcontrol}
 
Where the admissible pipeline is varied by resampling the data rather than by
changing a hyperparameter, each fit is explained on its own held-out sample, and
disagreement between fits confounds two sources: the fits differ, and the samples
they are explained on differ. The determinism check of
Section~\ref{method:regimes} excludes this by construction for a fixed split, and
cannot be applied here.
 
We separate the two sources without refitting, using the overlap between the
held-out sets. For a matched number of samples $m$, three conditions are
compared: attributions from one fit computed on two disjoint $m$-sample subsets
of its own held-out set; attributions from two fits computed on the same $m$
samples, drawn from the intersection of their held-out sets; and attributions
from two fits computed on disjoint $m$-sample subsets, the condition obtaining in
practice. Agreement is measured as the Jaccard overlap of the resulting top-$N$
sets.
 
Matching $m$ across conditions is necessary rather than cosmetic: attributions
estimated from fewer samples are noisier, and noise alone depresses agreement, so
an unmatched comparison would confound the effect being tested with the sample
size. The first condition also serves as the check that the measurement is not
noise-limited at the $m$ used, since it bounds how much of the observed
disagreement any sample-size effect could explain.
 
\section{Results}

\subsection{Equally accurate models do not share one explanation}
\label{sec:result1}

\begin{figure}[H]
    \centering
    \includegraphics[width=\linewidth]{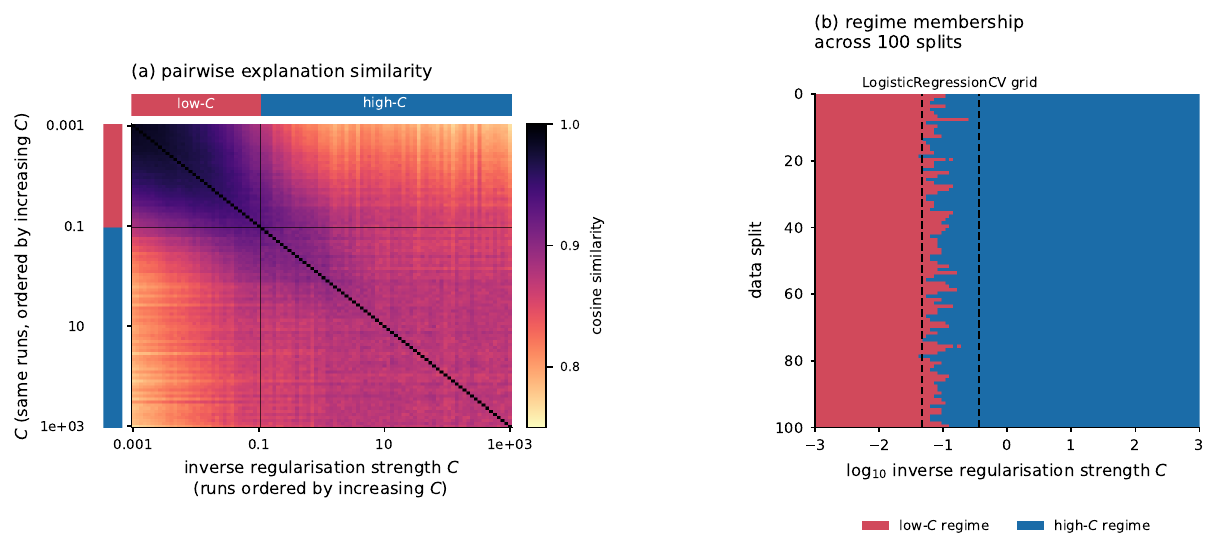}
    \caption{
        \textbf{Explanation similarity and regime membership along the
        regularisation path.}
        \textbf{(a)} Pairwise cosine similarity between explanation vectors
        for one example data split, with runs ordered by increasing $C$.
        Marginal colour strips indicate low-$C$ (red) and high-$C$ (blue)
        regime membership; black lines mark where $k$-means cuts the path.
        The colour scale spans cosine similarities from $0.75$ to $1$.
        \textbf{(b)} Regime membership across 100 data splits.
        Each row represents one split, with runs ordered by $C$.
        Dashed lines mark two points of \texttt{scikit-learn}'s default
        \texttt{LogisticRegressionCV} grid, $C \approx 0.046$ and
        $C \approx 0.36$, which fall in the low-$C$ regime in 95 and 0 of
        the 100 splits respectively.
        Here, $C$ denotes inverse regularisation strength.
    }
    \label{fig:population}
\end{figure}
We fitted L2-regularised logistic regression (lbfgs solver) to a TCGA pan-cancer
tumour-versus-normal task across a grid of 100 regularisation strengths
($C \in [10^{-3}, 10^{3}]$, log-spaced), repeated across 100 independent
train--test splits, giving $10^4$ model fits. Within each split every fit sees
the identical patient sample; only $C$ varies.

\paragraph{The population is not explanation-homogeneous.}
Explanation vectors within a split were more similar to one another within
regimes than between them: mean pairwise cosine similarity was $0.948$ (sd
$0.024$) within the low-$C$ regime and $0.883$ (sd $0.015$) within the high-$C$
regime, against $0.830$ (sd $0.045$) between them
(Figure~\ref{fig:population}a). The contrast is carried mainly by the low-$C$
regime: its within-regime similarity exceeds the between-regime value by $4.9$
standard deviations of its own spread and by $2.6$ of the between-regime spread,
whereas the high-$C$ regime is less tightly clustered and exceeds the
between-regime value by $0.053$, only $1.2$ of the between-regime spread. Both
regimes are internally coherent by the directional-coherence diagnostic of
Section~\ref{method:regimes}: $\rho = 0.975$ (sd $0.002$) in the low-$C$ regime
and $\rho = 0.940$ (sd $0.003$) in the high-$C$, against $\rho = 0.931$ for the
population taken as a whole. Coherence alone therefore does not set the regimes
apart from the whole: the population is concentrated overall, and all pairwise
similarities remain positive across the entire path. The structure is a graded
trajectory with distinguishable ends rather than two separated clouds, and we
use \emph{regime} in that sense: the two sides of the cut that $k$-means places
on it. What makes the ends worth distinguishing is not that they separate but
that, at matched accuracy, they report different genes (below and
Section~\ref{sec:result2}).

\paragraph{Content changes over a bounded stretch of the path.}
To locate where the change occurs, we placed each fit on the axis joining the
mean explanation of the ten lowest-$C$ fits in its split to that of the ten
highest-$C$ fits, so that progress runs from $0$ to $1$
(Section~\ref{method:regimes}). The change is neither uniform in $C$ nor
abrupt. Under LIME it is slowest at the low-$C$ end, fastest between
$C \approx 0.01$ and $0.1$, and largely complete by $C \approx 1$: $88\%$ of the
progress between the ends is made below that point, and over the remaining
three decades the explanation moves little further along the axis. The
10--90\% width of the change is $3.0$ of the six decades spanned (IQR
$2.5$--$3.2$), with its midpoint at $C \approx 0.057$ (IQR $0.051$--$0.067$).
SHAP gives the same shape, narrower and earlier: a width of $2.3$ decades, a
midpoint at $C \approx 0.023$, and $93\%$ of the change complete by
$C \approx 1$. For calibration, the same measure returns a width of about one
decade for a simulated sigmoid switch and about four for change spread evenly
across the range. The path is therefore a graded transition two to three
decades wide, followed by a high-$C$ plateau.

\paragraph{Membership follows the pipeline parameter, with patients fixed.}
Ordering the fits within a split by $C$, the $k$-means cut was a single
threshold in $98$ of $100$ splits; in each of the two exceptions (splits 819 and
875) a single run lies across it. A single threshold is what $k$-means returns
on any trajectory ordered by $C$, so this alone does not show that the regimes
are discrete; what it shows, with the progress measure above, is that
explanatory content is ordered by $C$ within every split. The cut fell between
$C \approx 0.040$ and $C \approx 0.248$ across splits, with a median of $0.081$
(Figure~\ref{fig:population}b), and in $91$ of $100$ splits it lies within a
quarter of a decade of the midpoint of the LIME progress curve: $k$-means
divides the path near the middle of the graded change. We use
\emph{transition} for that stretch of the path, not for a change of phase.

Because every fit within a split is trained on the same patients, this
observation excludes an entire class of alternative explanations by
construction rather than by statistical adjustment. Tumour type, histological
composition, tumour purity, class balance and the availability of adjacent
normal tissue are all constant within a split and therefore cannot account for
a difference between fits that varies only in $C$. Where a model sits on the
path is a property of the pipeline configuration.

\paragraph{Accuracy does not discriminate between the regimes.}
Across all $10^4$ fits, held-out accuracy was $0.9838 \pm 0.0027$. The regimes
are not exactly equal: the low-$C$ regime is more accurate by $0.21$ percentage
points on average, and in 96 of 100 splits when the full path is compared. The
difference is smallest near the transition. Restricting the comparison to models
within roughly a factor of five of the median $k$-means cut ($C \in [0.016, 0.4]$),
the difference is $0.145$ points, with the low-$C$ regime more accurate in 97 of
100 splits; restricted further to $C \in [0.04, 0.16]$ it falls to $0.055$
points---about a quarter of the within-split standard deviation of
$0.0021$---and the low-$C$ regime is more accurate in 56 of the 69 splits that
have fits on both sides of the transition within that window.

Models on either side of the transition therefore differ in predictive
performance by less than one percentage point, the tolerance used for the
headline results of \citet{marx2020predictive} and adopted throughout this
paper (Section~\ref{sec:alternatives}), while their explanations differ: the mean
between-regime cosine is $0.830$, and the top-50 genes of the lowest- and highest-$C$ profiles (each a mean over that position's fits) overlap at a Jaccard index of $0.505$. The low-$C$ regime
is consistently the more accurate, and we do not suppress that; the point is
that accuracy within the tolerance gives no reason to prefer one explanation
over the other. A practitioner selecting a single $C$ within the ordinary tuning
range gains a fifth of a percentage point by moving to the low-$C$ regime and
changes which genes the explanation prioritises, with no indication that the
second choice is being made at all.

\paragraph{The transition falls inside the range practitioners search.}
By default, \texttt{scikit-learn}'s \texttt{LogisticRegressionCV} searches ten
values of $C$ log-spaced between $10^{-4}$ and $10^{4}$ \citep{pedregosa2011scikit}.
Three of them lie near the transition: $C \approx 0.006$ falls in the low-$C$
regime in all 100 splits, $C \approx 0.046$ in 95, and $C \approx 0.36$ in
none, while the \texttt{LogisticRegression} default of $C = 1$ is high-$C$ in
every split. The library's default search therefore spans both regimes in every
split, and three of its points lie on the graded part of the path, where content
is still changing, while the default $C = 1$ sits where the plateau begins. The
regime of the selected model is set by which grid point cross-validation
returns, not by a choice the practitioner makes. A fixed, documented value can
sit close to the cut. $C = 0.1$, selected by cross-validation for a deployed
text classifier \citep{bernardini2021} and used in a published marker-gene
method \citep{liu2026ilr}, falls on the low-$C$ side of the cut in 26 of the
100 splits and on the high-$C$ side in the other 74: the same setting, on the
same data, sits on a different side of the transition depending on the
train--test split.
\subsection{Explanatory content varies systematically along the path}
\label{sec:result2}
\begin{figure}[H]
    \centering
    \includegraphics[width=0.5\linewidth]{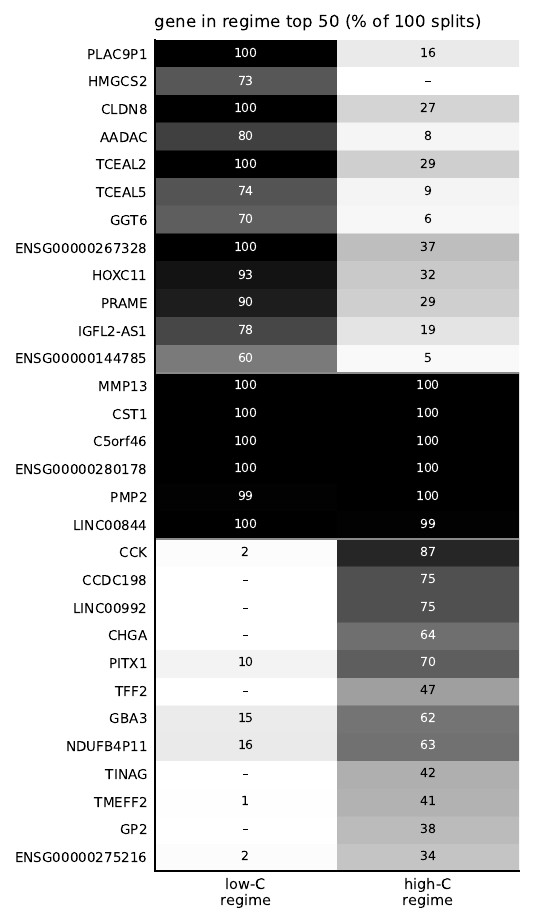}
        \caption{\textbf{Genes that distinguish the regimes.} Percentage of 100
splits in which each gene enters the top 50 of the low-$C$ and the high-$C$
regime profile (LIME; a dash marks zero). Rows are chosen by rule, not by hand:
the twelve genes with the largest recurrence difference in favour of each
regime, and the six genes most recurrent in both. Profiles are computed per
split, with no averaging across splits. Recurrence in a profile is a property of
the models, not a significance claim about the biology. Unannotated loci are
shown by Ensembl identifier.}
    \label{fig:regimes}
\end{figure}
\begin{table}[t]
\centering
\small
\caption{The two attribution lenses resolve the same 100 model fits per split at
different granularity. Regions are ordered by mean $C$. LIME returns
$k^\star = 2$ in all 100 splits. SHAP's $k^\star$ varies across splits (2 in 19,
3 in 56, 4 in 18, 5 in 4, 6 in 3), and its silhouette profile is nearly flat
from $k = 3$ to $k = 5$ (Table~\ref{tab:silhouette}); the SHAP rows describe the
56 splits with $k^\star = 3$, whose regions are contiguous in $C$ in 48. SHAP's
lowest-$C$ region lies inside LIME's low-$C$ regime and its highest-$C$ region
inside LIME's high-$C$ regime, but its intermediate region straddles the LIME
boundary (on average 11 of its 30 runs on the low-$C$ side), so the two
partitions are not nested. Runs are means over splits; $C$ ranges are those of
split 800, and the LIME boundary is the median over splits. The silhouette
column gives the mean over all 100 splits at that $k$.}
\label{tab:partitions}
\begin{tabular}{@{}llrlr@{}}
\toprule
Lens & Region (by mean $C$) & Runs & Approx.\ $C$ range & Silhouette at $k$ \\
\midrule
LIME ($k^\star = 2$, 100/100 splits)
 & low-$C$ & 32 & $10^{-3}$ -- $0.081$ & \multirow{2}{*}{0.252} \\
 & high-$C$  & 68 & $0.081$ -- $10^{3}$  & \\
\midrule
SHAP ($k^\star = 3$, 56/100 splits)
 & lowest $C$ & 22.0 & $10^{-3}$ -- $0.019$ & \multirow{3}{*}{0.665} \\
 & intermediate  & 30.1 & $0.022$ -- $1.2$    & \\
 & high-$C$         & 47.8 & $1.4$ -- $10^{3}$   & \\
\bottomrule
\end{tabular}
\end{table}

A difference in attribution vectors is not by itself a difference in
explanation. To establish that the population differs in a way a practitioner
would act on, we carried each region's profile through the analysis a
practitioner would perform: over-representation analysis of its top-ranked
genes, per split and per region, with no averaging of profiles across splits.

\paragraph{The number of regions is lens-dependent; the grid dependence is not.}
The two attribution lenses partition the same 100 fits differently. LIME
returns $k^\star = 2$ in all 100 splits, and its silhouette profile peaks at
$k = 2$ ($0.252$) in every split and declines thereafter. SHAP does not return
a single count: $k^\star$ is 3 in 56 splits, 2 in 19 and 4 or more in 25, and
its mean silhouette profile is nearly flat from $k = 3$ to $k = 5$ ($0.665$,
$0.662$, $0.651$; Table~\ref{tab:silhouette}), with the maximum at $k = 3$ in
only half the splits. By the criterion of Section~\ref{method:regimes}, that is
the profile of a connected structure that $k$-means segments at a granularity
the data do not fix, rather than of a resolved partition.

Where SHAP does return three regions they are contiguous in $C$ in 48 of 56
splits, and its lowest- and highest-$C$ regions lie within LIME's two regimes
(Table~\ref{tab:partitions}). Its intermediate region straddles the LIME
boundary, however, so the partitions are not nested (cluster purity $0.889$),
and we do not treat either as the correct one. What the pipeline determines is
how explanatory content varies with $C$; how many regions a given lens resolves
in that variation is a property of the lens, and under SHAP it is not stable
even across splits. By the criteria of Section~\ref{method:regimes} the number of
regions is therefore lens-dependent---LIME resolves two, SHAP's profile is flat
from $k = 3$ to $5$---and we abstain from claiming a number. What both lenses
agree on, and what the remaining analyses use, is that explanatory content varies
reproducibly with position along the path (Section~\ref{method:positions}).
This is not only a caution. The aggregation analyses of
Sections~\ref{sec:result3} and~\ref{sec:stability} are computed on a
decomposition of the path by $C$ alone, with no clustering entering the
measurement, and the central quantity of Section~\ref{sec:result3}---which
fitted model a practitioner's reported panel corresponds to, given where their
grid sits---involves no decomposition at all.
\paragraph{What the feature space already contains.}
Enrichment compares a gene list against a background, and the choice of
background determines what can be significant. The modelling feature space here
is the 1{,}000 genes with the highest variance across samples, selected from
the transcriptome before any split is drawn or any model is fitted, and that
selection is not neutral with respect to the ontology. Testing the feature
space itself against the genome-wide domain, it is significantly enriched for
184 terms. Extracellular region occupies 335 of the 1{,}000 genes
($33.5\%$, $p = 9 \times 10^{-90}$), and the epithelial terms are
over-represented as well---formation of the cornified envelope, keratinisation
and epithelium development among them ($36$, $22$ and $70$ genes). The metabolic
and secretory terms are not: PPAR signalling, butanoate metabolism, acyl-CoA
ligase activity, neuropeptide receptor binding and the secreted cystatins are
each non-significant in the feature space.

The most concentrated signature in the feature space is immunological:
immunoglobulin complex contains 110 of the 1{,}000 genes
($p = 2 \times 10^{-126}$), and eleven of the twelve most significant terms are
immune-related. A variance-based selection on a tumour-versus-normal contrast
favours genes that differ strongly between the classes, and immune infiltration
is among the largest such differences. The feature space therefore carries
structure of its own, independent of any model fitted within it. Tested against
the genome-wide domain, a region's top-50 genes would inherit that structure,
and terms recurring in a region profile could reflect the pre-filter rather than
the models.

We therefore test the region profiles against the feature space itself, the set
from which every profile is drawn. A term already concentrated in the feature
space is not thereby barred from significance against it---the test asks
whether the query is enriched \emph{relative to} that domain---but the headroom
is small, and a 50-gene query rarely clears it.

\paragraph{The regimes differ in genes, not in enriched pathways.}
We use three positions throughout, defined by thresholds in $C$ rather than by
any clustering: \emph{lowest-$C$} ($C \le 0.016$), \emph{intermediate}
($0.016 < C \le 0.081$) and \emph{high-$C$} ($C > 0.081$). The upper boundary is
the median LIME transition reported above and the lower lies near the lower SHAP
boundary ($C \approx 0.019$ in the splits where SHAP resolves three regions), so
both are estimated rather than arbitrary; what the decomposition avoids is not
the estimate but the dependence on a partition procedure whose granularity is
lens-specific, and Section~\ref{sec:alternatives} repeats the analyses with round
boundaries.

At gene level the regimes differ reproducibly (Figure~\ref{fig:regimes}).
Four low-$C$-specific genes enter the low-$C$ profile's top 50 in all 100 splits
(\textit{PLAC9P1}, \textit{CLDN8}, \textit{TCEAL2} and one unannotated locus)
and the high-$C$ profile's in at most 37; \textit{HMGCS2} enters the low-$C$
top 50 in 73 splits and the high-$C$ in none. No high-$C$-specific gene is as
reproducible: the most recurrent, \textit{CCK}, enters the high-$C$ top 50 in
87 splits and the low-$C$ in 2, and six of the twelve (\textit{CCDC198},
\textit{LINC00992}, \textit{CHGA}, \textit{TFF2}, \textit{TINAG} and
\textit{GP2}) never enter the low-$C$ top 50. Six genes are shared, entering
both profiles in at least 99 of 100 splits. The asymmetry is the gene-level face
of the difference in reproducible cores reported below (Table~\ref{tab:cores}).
These differences do not assemble into distinct enriched pathways. Against the
feature-space background, one term reaches significance in one of the 200
regime panels (100 splits, two regimes), and none in the other 199. That null
is informative only if the test could have detected a real difference, so we
measured its power by planting gene sets of known size into regime panels: a
30-gene immunoglobulin set was detected at $p \approx 10^{-14}$, whereas a
planted 8-gene keratinisation set was not detected, and a panel of 50 random
feature-space genes returned no term. A 50-gene panel tested
against a 1{,}000-gene background can therefore detect a large coherent set and
cannot detect a small one. A competitive gene-set test on the two LIME regimes, which uses the whole
attribution profile rather than a top-50 cut, finds no gene set that separates the regimes in a majority of
splits. The sets most often weighted towards the low-$C$ regime are broad
(response to nutrient levels and intracellular signal transduction, each in 35 of
100 splits), with lipid metabolic process in 26 and lipid catabolic process in
20; towards the high-$C$ regime, RNA processing appears in 20 and no other set in
more than 6. The test's null treats genes as exchangeable and ignores their
correlation, so these counts are if anything generous. We therefore describe the regimes by their genes throughout, and make no claim
that they implement different biological programmes. Genome-wide enrichment of
the same profiles returns many terms, including the epithelial terms the
pre-filter itself carries, and we do not report those as findings.

\paragraph{Reproducibility differs between positions.}
Whether a position's content can survive any summary depends on whether it
recurs at all. Taking each position's profile per split and recording the genes
that enter its top 50 in at least 70\% of the 100 splits---its reproducible
core---the three positions differ, and not in the direction the run counts
would suggest (Table~\ref{tab:cores}). The high-$C$ position supplies 68 of
every 100 fits and has the \emph{smallest} core, 30 genes of which 4 are
exclusive to it under LIME; the lowest-$C$ position supplies 20 fits and has the
largest, 45 genes of which 11 are exclusive. SHAP gives the same ordering on the
same decomposition. A practitioner sweeping this grid therefore spends the
majority of their fits at the position whose explanatory content is least
reproducible, and a minority at the position where it is most.
Section~\ref{sec:result3} shows what that allocation costs the reported panel.
 
\begin{table}[t]
\centering
\small
\caption{Reproducibility by position on the path. The \emph{core} of a position
is the set of genes entering its profile's top 50 in at least 70\% of the 100
splits; \emph{exclusive} counts those in no other position's core. Positions are
defined by $C$ alone (Section~\ref{sec:result2}), so the same decomposition
applies under both lenses. The high-$C$ position contributes the most fits and the
least reproducible content.}
\label{tab:cores}
\begin{tabular}{@{}lrrrrr@{}}
\toprule
& & \multicolumn{2}{c}{Core} & \multicolumn{2}{c}{Exclusive} \\
\cmidrule(lr){3-4}\cmidrule(l){5-6}
Position & Fits per split & LIME & SHAP & LIME & SHAP \\
\midrule
Lowest $C$ ($C \le 0.016$)      & 20 & 45 & 35 & 11 & 7 \\
Intermediate ($0.016 < C \le 0.081$) & 12 & 37 & 28 & \phantom{0}2 & 0 \\
High $C$ ($C > 0.081$)                     & 68 & 30 & 27 & \phantom{0}4 & 4 \\
\midrule
Common to all three                     & -- & 25 & 21 & -- & -- \\
\bottomrule
\end{tabular}
\end{table}

\paragraph{Independent support is uneven.}
The genes that distinguish the regimes are the candidates a practitioner would
carry forward, so we record whether their tumour--normal direction is supported
outside this cohort, prioritising protein-level studies over transcriptomic
reanalyses of the same data (Table~\ref{tab:literature}). Among the
low-$C$-specific genes, \textit{HMGCS2} is reduced in colorectal tumour relative
to normal mucosa by immunohistochemistry \citep{camarero2006hmgcs2,hmgcs2crc2019};
\textit{PRAME} protein is confined among normal tissues to testis and
proliferative endometrium and is expressed across a wide range of tumours
\citep{kaczorowski2022prame}; \textit{HOXC11} is higher in colon adenocarcinoma
than in normal tissue at mRNA and protein level \citep{liu2021hoxc11};
\textit{TCEAL2} is lower in clear-cell renal carcinoma than in adjacent kidney by
immunohistochemistry and RT-qPCR \citep{Zhou2020TCEAL2}; and \textit{GGT6} is
lower in colorectal tumour on tissue microarrays \citep{Yang2026GGT6}.
\textit{CLDN8} is conflicting: higher in tumour at protein level in one colorectal
series of 20 matched pairs \citep{cheng2019cldn8} and lower at RNA level in 30
microdissected pairs \citep{grone2007claudin}. \textit{AADAC} and
\textit{IGFL2-AS1} have tissue-dependent or RNA-only evidence
\citep{Wang2023AADAC,AbdelTawab2022AADAC,Qin2023IGFL2AS1,Marhamati2026IGFL2AS1}.
On the high-$C$ side, \textit{CHGA} is lower in prostate adenocarcinoma than in
adjacent benign epithelium \citep{SionVardy2004CHGA}, while \textit{TFF2} is
lower in gastric cancer \citep{Jiang2014TFF2,Jung2010TFF2} but higher in
colorectal cancer \citep{Yu2015TFF2}, and \textit{PITX1} lower in gastric cancer
by immunohistochemistry \citep{Chen2008PITX1} but higher in clear-cell renal
carcinoma at RNA level \citep{Zhang2021PITX1}; for \textit{CCK} we found no
matched tumour--normal comparison. No independent evidence was found for the
remaining genes, several of which are pseudogenes or unannotated loci. Every gene
with protein-level evidence in a single direction---\textit{HMGCS2},
\textit{PRAME}, \textit{HOXC11}, \textit{TCEAL2}, \textit{GGT6} and
\textit{CHGA}---agrees with its within-site direction in this cohort
(Table~\ref{tab:tissue}). The numbers are small, and we read this as consistency
rather than confirmation. Corroboration of this kind establishes a gene as a
tumour--normal marker; it does not establish that the regime recruiting it is
biologically preferable, and we describe recurrent genes as reproducible model
interpretations throughout.

\paragraph{Changes in patient composition do not account for the within-split dependence.}
The genes that distinguish the regimes are tissue-restricted and TCGA's
adjacent-normal samples are unevenly distributed across cancer types, so
different positions on the path might weight different normal-tissue contrasts
rather than different tumour biology. Patient membership cannot vary within a
split, so differences between fits in one split cannot come from which patients
are included; regularisation could still change how much weight the models give
to tissue or composition signals present in those fixed patients. We therefore
also examined the genes directly. For each gene we computed the
tumour-versus-normal effect size within each tissue source site with at least
eight samples per class (30 sites), for each of the 24 regime-specific genes of
Figure~\ref{fig:regimes} (Table~\ref{tab:tissue}).
The two
regimes' genes behave differently. Of the twelve low-$C$-specific genes, ten
separate tumour from normal in the same direction in a majority of sites
(at least 16 of 30), and none separates in the opposite direction in more than
six: \textit{TCEAL2} is lower in tumour in 29 of 30 sites and higher in none,
\textit{HMGCS2} lower in 24 and higher in 2. Of the twelve
high-$C$-specific genes, five reach such a majority, while \textit{CCDC198},
\textit{LINC00992}, \textit{TFF2}, \textit{TINAG} and \textit{GP2} separate in
opposite directions in at least six sites each---the behaviour expected of
tissue-restricted genes whose tumour--normal contrast depends on the tissue
(\textit{GP2}, for instance, reaches $d = -10.2$ in one site).
The low-$C$
genes therefore behave as broadly consistent tumour--normal markers, and the
high-$C$ genes more often as markers of tissue identity. These within-site
associations support the genes' tumour--normal relevance across sites; they do not
establish independence from cell composition, which we leave unresolved
(Appendix~\ref{app:alternatives}).

\subsection{The conventional summary reweights the population by the grid}
\label{sec:result3}
\begin{figure}[H]
    \centering
    \includegraphics[width=0.75\linewidth]{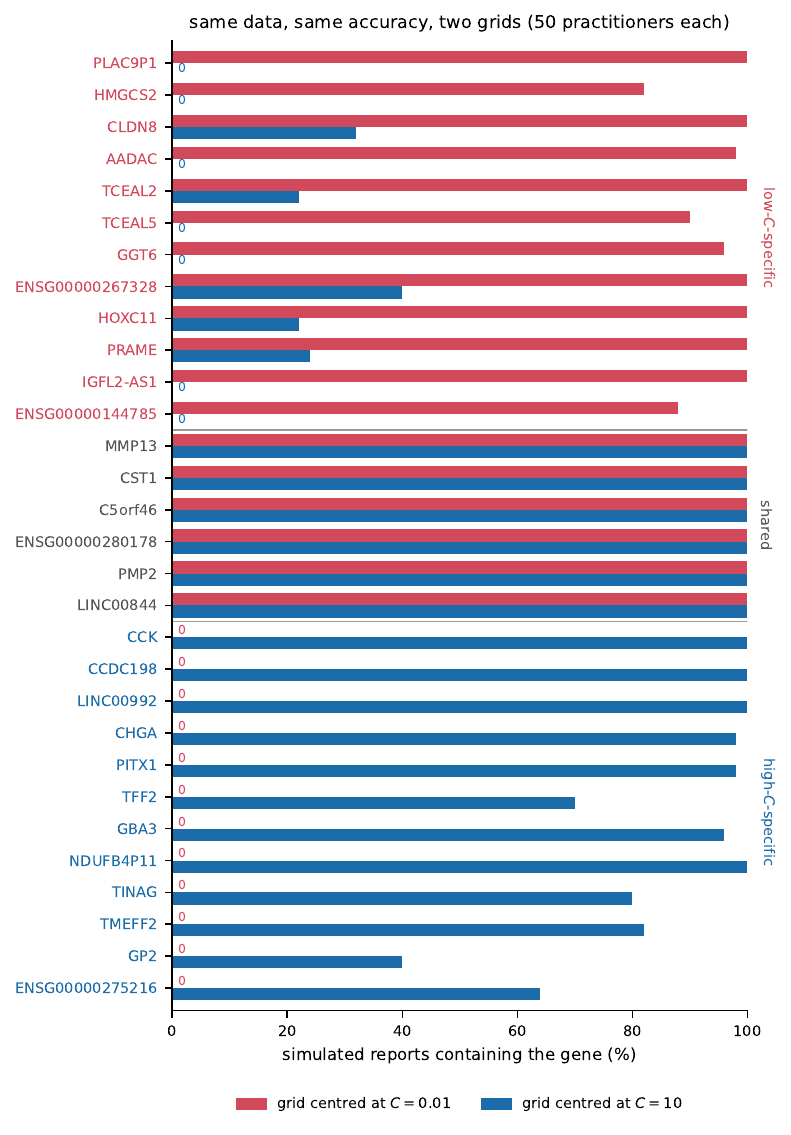}
    \caption{\textbf{The grid decides which genes are reported.} Fifty
simulated practitioners each draw ten data splits and report the top 50 genes
of their averaged explanation, once under a ten-point grid centred at $C = 0.01$
and once under a grid centred at $C = 10$ (each spanning $\pm 1.5$ decades),
with the same splits drawn under both. Bars give the percentage of reports
containing each gene. Rows are the genes of Figure~\ref{fig:regimes}, grouped as
specific to the low-$C$ regime, shared, and specific to the high-$C$ regime.
The twelve high-$C$-specific genes appear in no report from the low-centred
grid; seven of the twelve low-$C$-specific genes appear in no report from the
high-centred grid; the shared genes appear in every report from both. Mean
held-out accuracy differs between the two grids by $0.06$ percentage points.
Unannotated loci are shown by Ensembl identifier.}
    \label{fig:aggregation}
\end{figure}

\paragraph{Simulating the reported explanation.}
We simulated 50 independent practitioners, each following the workflow
described in Section~\ref{method:phantom}: draw ten data splits, scan a ten-point
log-spaced grid of $C \in [0.01, 10]$ within each, and average the resulting
unit-normalised explanation vectors into a single reported vector, whose top 50
genes form the reported panel. Each practitioner therefore aggregates 100
model fits. Gene-level results in this subsection are computed on LIME
attributions; the positional results that follow are computed under both
lenses, and the claim that the finding does not depend on the attribution method
rests on those. The grid is narrower than \texttt{scikit-learn}'s
\texttt{LogisticRegressionCV} default and spans the transition, as a
practitioner's grid would if they did not know a transition existed.

\paragraph{The grid decides which genes are reported.}
Averaging does not report the population; it reports the part of it the grid
samples. Holding the practitioners and their drawn splits fixed and changing
only the grid's centre, from $C = 0.01$ to $C = 10$, divides the reported panels
by regime (Figure~\ref{fig:aggregation}). The low-centred grid places
77\% of its fits in the low-$C$ regime (70--90\% across splits) and the
high-centred grid all of them in the high-$C$ regime. The twelve
genes specific to the high-$C$ regime appear in none of the 50 reports from the
low-centred grid and in roughly 40--100\% of those from the high-centred
grid. Seven of the twelve low-$C$-specific genes, including \textit{HMGCS2},
appear in no report from the high-centred grid, against more than 80\% of
reports from the low-centred one. The genes both regimes share appear in every
report from both grids. The two grids' mean held-out accuracies differ by $0.06$
percentage points. Nothing in either report indicates that the other exists.
\paragraph{Aggregation keeps what is shared, and the rest in proportion to the grid.}
At first sight the reported panel loses little. Defining each region's signature
as the genes appearing in its profile in at least 70\% of splits (25 shared, 18
low-$C$-specific, 4 high-$C$-specific), the panel recovers the shared component
completely and most of both region-specific sets: $68\%$ of the low-$C$-specific
signature and $88\%$ of the high-$C$-specific one (Table~\ref{tab:sigexcl}). This
excludes the account on which aggregation reports only the intersection of the
two regimes. It does not mean the population is reported. Measured against the
content \emph{exclusive} to each end of the path---the genes no other position
carries reproducibly---the same panels recover $88\%$ of the high-$C$ set but
only $25\%$ of the low-$C$ one, and both figures are set by the grid rather than
by the data: a grid confined to the high-$C$ side recovers $13\%$ of the low-$C$
exclusive core, and one confined to the low-$C$ end recovers none of the
high-$C$ set. What aggregation preserves is the content every model shares;
what it loses is what distinguishes the end of the path the grid samples least,
and nothing in the panel shows which end that was.

Nor is the panel detached from the models it summarises. Its panel provenance
(Section~\ref{method:provenance})---the fraction of its 50 genes found in the
top 50 of its single best-matching constituent model---is $0.820$ under LIME
(sd $0.020$) and $0.843$ under SHAP (sd $0.023$): 41 or 42 of its 50 genes sit
inside one fitted model, and no sampled fit contains all 50. How that compares
with the agreement between individual models depends on the lens. Under LIME,
two models from the same regime share $0.784$ (low-$C$) and $0.620$ (high-$C$)
of their top 50, below the panel's figure; under SHAP, which is deterministic
for logistic regression and carries none of LIME's sampling variance, they share
$0.928$ and $0.954$, above it. The panel is therefore largely contained in one
real model but less similar to it than models of the same regime are to one
another, and the practitioner did not choose that model: under both lenses it
lies at a median $C \approx 0.0215$, more than a decade below the centre of their grid.

\paragraph{The panel corresponds to a position on the path, set by the grid.}
It does not correspond to a model the practitioner selected. The best-matching
constituent lies at a median $C \approx 0.0215$ (interquartile range
$0.0093$--$0.043$), more than a decade below the grid's geometric centre
($0.316$) and below the median transition ($0.081$); 30\% of best matches fall
inside the window $[0.040, 0.248]$ in which the transition occurs. The landing
point is a property of the population rather than of the averaging arithmetic:
repeating the aggregation under four rules differing only in per-vector
scaling---$\ell_2$, $\ell_1$, division by the largest component, and rank
averaging, in which every model contributes equally---leaves the median
best-matching model at $C \approx 0.0215$ under all four rules and both lenses,
and the effective support of the attribution vectors changes modestly along the
path: under SHAP the participation ratio rises from 266 to 366 genes across six
decades of $C$, and under LIME it is flat at 604--620.

What determines it is where the grid sits. Sliding a ten-point grid of fixed
width along the path and repeating the simulation at each position
(Table~\ref{tab:gridcentre}), the best-matching model tracks the grid centre.
Two correlations describe this and we report both, because they answer different
questions. Across the seven grid centres, the correlation in $\log C$ between
centre and the \emph{median} best match at that centre is $0.960$; across all
350 individual practitioners it is $0.88$ (Spearman $0.92$). The first
summarises the table; the second is what it means to say a practitioner's grid
predicts their match, and is the figure to weigh. The dependence is monotone but
not linear. At centres $0.316$ and $1$ the best match stays near the grid's
low-$C$ end ($C \approx 0.02$--$0.03$, below the transition) although most of
the grid lies above it, and it moves to the high-$C$ side only once the centre
reaches $3$: averaging holds the panel on the low-$C$ side of the path longer
than the grid's own weighting would suggest. The content each position uniquely
carries is reported in full when the grid sits on that position and largely not
otherwise. The same $10^4$ fits and the same averaging rule produce a panel
built on the most heavily regularised models or on the least, according to a
choice usually reported, if at all, as a tuning detail.

The result does not depend on the attribution lens. Repeating the positional
analysis under SHAP---which resolves no stable partition at all, so that no
binary regime partition is available---the best-matching model tracks the grid
centre with a correlation in $\log C$ of $0.89$ between centre and median match
across the seven centres, and $0.84$ (Spearman $0.89$) across the 350
individual practitioners, and recovery of the content
exclusive to the lowest-$C$ position falls from $99\%$ to $46\%$ across the
same range of grid centres. The positional analysis is defined by thresholds in
$C$ and requires no partition, which is what allows it to be computed
identically under both lenses.

\begin{table}[t]
\centering
\small
\caption{The reported panel follows the grid. A ten-point log-spaced grid of
fixed width ($\pm 1.5$ decades) is slid along the regularisation path and the
practitioner simulation repeated at each position over the same $10^4$ fits.
``Best-match $C$'' is the regularisation strength of the constituent model
whose own top-50 most closely matches the reported panel, as the median over 50
practitioners at each centre. The last two columns give recovery of the gene
content \emph{exclusive} to the lowest-$C$ position (11 genes under LIME) and to
the high-$C$ position (4 genes), each defined by recurrence in at least 70\% of
splits; the high-$C$ column rests on four genes and should be read with that
count in view, and the intermediate position is omitted because only two genes
are exclusive to it. The correlation in $\log C$ between grid centre and the
median match is $0.960$ over the seven rows, and between grid centre and each
individual practitioner's match it is $0.88$ (Spearman $0.92$, $n = 350$).}
\label{tab:gridcentre}
\begin{tabular}{@{}lllrr@{}}
\toprule
Grid centre & Realised $C$ range & Best-match $C$
  & \multicolumn{1}{c}{Lowest-$C$ (11)} & \multicolumn{1}{c}{High-$C$ (4)} \\
\midrule
$0.01$  & $0.001$--$0.305$  & $0.0015$ & 86\% & 0\%   \\
$0.03$  & $0.001$--$0.933$  & $0.0046$ & 67\% & 12\%  \\
$0.10$  & $0.00305$--$3.27$ & $0.0071$ & 43\% & 59\%  \\
$0.316$ & $0.00933$--$10$   & $0.0215$ & 27\% & 87\%  \\
$1.0$   & $0.0327$--$30.5$  & $0.0327$ & 19\% & 100\% \\
$3.0$   & $0.100$--$93.3$   & $0.6984$ & 17\% & 100\% \\
$10$    & $0.305$--$327$    & $3.2745$ & 16\% & 100\% \\
\bottomrule
\end{tabular}
\end{table}

\paragraph{Content is lost by under-sampling, and returns when sampled.}
Averaging retains what the grid samples, in proportion to how heavily it
samples it, and the content exclusive to either end of the path behaves the same
way (Table~\ref{tab:gridcentre}). The four genes exclusive to the high-$C$
position appear in no panel when the grid is centred at $C = 0.01$ and in every
panel once the centre reaches $C = 1$; the eleven exclusive to the lowest-$C$
position are recovered at $86\%$ when the grid sits on them and at $16$--$27\%$
once it is centred at $0.316$ or above. A ten-point grid over $[0.01, 10]$
reaches the lowest-$C$ position at a single point, so nine of every ten
constituents attribute that content at the lower rate found elsewhere on the
path, and it is outvoted. We found no position whose exclusive content fails to
survive a grid that samples it densely: in this pipeline the loss is a property
of the sweep rather than of the content, and moving the grid recovers it.

The genes follow the same pattern individually. \textit{HMGCS2}, which enters the
low-$C$ top 50 in 73 splits and the high-$C$ in none, ranks at a median of 113 in the averaged vector
of the conventional grid, against 44 in its own region's profile, and appears in
none of the 50 panels; of the 18 low-$C$-specific signature genes, 6 are retained
by every panel and three by two panels or none.

The asymmetry has a simple arithmetic, which requires distinguishing two
sampling distributions that are easy to conflate. The full 100-point path over
$C \in [10^{-3}, 10^{3}]$ places 20, 12 and 68 of its fits in the three
positions, and the reproducible cores of Table~\ref{tab:cores} are computed on
that path. A practitioner does not sample that path: the ten-point grid over
$C \in [0.01, 10]$ places one grid point in the lowest-$C$ position, two in
the intermediate and seven in the high-$C$, so across ten splits their 100 fits
divide 10, 20 and 70.

\begin{table}[htbp]
\centering
\small
\caption{Signature recovery and exclusive-core recovery, measured on the same
reported panels. ``Signature'' is the region-specific set from the two-region
partition of Section~\ref{sec:result1}: 18 low-$C$-specific and 4 high-$C$-specific
genes, each entering that region's profile in at least $70\%$ of splits.
``Exclusive core'' is the set exclusive to that position in the three-position
decomposition of the path: 11 and 4 genes (Table~\ref{tab:cores}); on the
high-$C$ side the two reference sets coincide. Each entry
is the mean over 50 simulated practitioners, each drawing 10 splits and a
ten-point log-spaced grid within the span; standard deviation in brackets. The
25-gene shared component is recovered completely under every span and is
omitted. This table is computed by a single script over both reference sets, and
is the source for every signature- and span-based recovery percentage quoted
in the text. Its
lowest-$C$ figure for the grid used here ($25\%$) and the corresponding entry
of Table~\ref{tab:gridcentre} ($27\%$ at grid centre $0.316$) measure the same
quantity by two independent constructions, and agree to within two points. Both use
the aggregation rule of Section~\ref{method:phantom}: signed attribution
vectors are averaged and the result ranked by magnitude, so that a gene whose
sign varies across the fits averaged can cancel. The grid over $[0.001, 0.1]$
draws 92\% of its fits from the low-$C$ regime, since its top point lies above
the transition in some splits.}
\label{tab:sigexcl}
\begin{tabular}{@{}lcccc@{}}
\toprule
& \multicolumn{2}{c}{Region signature} & \multicolumn{2}{c}{Position-exclusive core} \\
\cmidrule(lr){2-3}\cmidrule(lr){4-5}
Grid span & low-$C$ (18) & high-$C$ (4) & low-$C$ (11) & high-$C$ (4) \\
\midrule
$[0.01, 10]$, used here             & 68 (6)  & 88 (12) & 25 (7) & 88 (12) \\
$[0.05, 1]$                          & 63 (7)  & 89 (13) & 21 (5) & 89 (13) \\
\texttt{LogisticRegressionCV} default & 73 (5)  & 76 (18) & 39 (9) & 76 (18) \\
$[0.001, 0.1]$, low-$C$ end          & 100 (1) & 0 (0)   & 87 (6) & 0 (0) \\
$[1, 10^{3}]$, high-$C$ only            & 36 (6)  & 100 (0) & 13 (6) & 100 (0) \\
\bottomrule
\end{tabular}
\end{table}

The asymmetry between signature and exclusive-core recovery
(Table~\ref{tab:sigexcl}) follows the grid's own weighting: the high-$C$
position receives seven of the ten grid points---70 of every 100 fits a
practitioner produces---and its exclusive content survives at $88\%$, while the
lowest-$C$ position receives a single grid point, 10 fits in 100, and its
exclusive content is recovered at $25\%$. Recovery rises with the weight the
grid places on a position and exceeds that weight in both cases, as a top-50 cut
rewards being sampled at all. Read down the low-$C$-exclusive column and the
result is Table~\ref{tab:gridcentre} measured on genes rather than on model
identity---$87\%$ when the grid sits at the low-$C$ end, $21$ to $39\%$ under the
straddling spans, $13\%$ when the grid sits entirely beyond it. The two
measurements differ because a region's signature includes everything that region
expresses reproducibly, most of which the other region also expresses at some
level, so it survives an uneven sample; the exclusive core is by construction the
part no other position carries reproducibly.

\paragraph{The grid's centre decides what a researcher would investigate.}
The measurements above concern what the reported panel contains. Whether that
difference reaches a decision we tested with a procedure fixed before the
comparison was run. Fifty simulated researchers each draw ten splits and sweep
two grids over \emph{the same} splits, so that the only difference between their
two reports is the grid. Each builds a 50-gene panel as above and carries the
top ten candidates forward for follow-up, under two rules declared in advance:
the top ten outright, and the top ten after removing the 25-gene shared
component, which a researcher would treat as known biology.

Between grids centred three decades apart---at $C = 0.01$ and at $C = 10$, each
spanning $\pm 1.5$ decades---the shortlists share $6.8$ of ten candidates under
the first rule and $2.4$ under the second. Once known biology is set aside,
about three of every four genes a researcher would carry to the bench differ,
from identical data, identical splits and mean held-out accuracies $0.06$
percentage points apart. What matters is the centre, not the width: grids that
both straddle the transition and differ mainly in width---\texttt{scikit-learn}'s
\texttt{LogisticRegressionCV} span, eight decades centred at $C = 1$, and the
$C \in [0.01, 10]$ grid used here, three decades centred at $C \approx 0.32$,
both clipped to the fitted path---share $9.1$ and $7.9$ of ten, at accuracies
$0.22$ points apart. Both accuracy differences lie inside the tolerance of
Section~\ref{sec:alternatives}, and a second draw of fifty practitioners
reproduces all four figures to within a fifth of a gene ($6.6$, $2.5$, $9.0$ and
$7.8$). This is the asymmetry of Table~\ref{tab:sigexcl}, carried from content
to decision.

The result concerns allocation, not outcome: a fixed prioritisation procedure,
given a fixed budget and panels from two defensible grids at equivalent
accuracy, spends that budget on different genes. It does not show which
shortlist is correct, that researchers prioritise in exactly this way, or that
any experiment would succeed or fail.

\paragraph{The reported combination is composed by the aggregation step.}
Every gene in a reported panel is supported by some fitted model; the
combination is not. Scoring each panel by its joint coverage of the two region
signatures (Section~\ref{method:provenance}), 43 of the 50 simulated
practitioners report a panel holding more of both at once than any of the 100
models they averaged (median $67\%$, against $56\%$ for the best of a
practitioner's own constituents). The combination exists in the wider
population---the best of all $10^4$ fits reaches $75\%$---but not among the
models the practitioner trained. The reported explanation is therefore not the
explanation of any model that was fitted, validated on the task, or could be
deployed: accuracy was established for each constituent, and for none of them
is the panel the explanation. Nor is it the explanation of an ensemble, which
the workflow neither builds nor evaluates, and which, after per-vector
normalisation, the average would not describe exactly in any case. A claim
traced back from the panel reaches the averaging operation rather than a
validated model, and the traceability chain of Section~\ref{sec:related} breaks
at that step.

\paragraph{Cross-model-class consensus does not recover what is lost.}
A common response to explanation instability is to widen the pool: fit several
model classes and retain the features that recur across all of them. We tested
this directly by pooling every individual fit from the regularisation path, a
seed-varied deep network and a seed-varied gradient-boosted ensemble
(Section~\ref{sec:scope}), 10{,}000 fits from each, and applying
frequency-thresholded selection across the combined population.

Pooling does not recover the regime-specific content; it removes more of it
(Table~\ref{tab:crossclass}). Coverage is measured against the genes appearing in
a LIME regime profile of the regularisation path in at least 90\% of splits: 16
shared, 16 specific to the low-$C$ regime and 2 to the high-$C$ regime. At a
stability threshold of 70\%, the boosted population alone retains 13 of the 16
low-$C$-specific genes and the regularisation path alone both high-$C$-specific
genes; the pooled panel of 19 genes retains 6 and none. At 90\% the pooled panel
is 8 genes, all from the shared set, while the boosted population alone still
retains 11 of the 16 low-$C$-specific genes. The pooled panel is never larger,
and never recovers more of either regime's genes, than the best single population
at the same threshold, and no high-$C$-specific gene survives pooling at 70\% or
above.

In these populations and at these thresholds, pooling did not help. A
frequency threshold retains what most models agree on, and here adding
heterogeneous model classes lowered the selection frequency of the genes that
distinguish positions. Pooling need not always shrink a panel, since it can
raise some genes' frequencies as well as lower others'.
\subsection{Selection by recurrence trades content for provenance}
\label{sec:stability}

The preceding results concern averaging. Practice, however, frequently does not
average: ensemble feature selection, consensus nested cross-validation and
frequency-thresholded selection all retain the features that \emph{recur}
across models and report those as the finding
\citep{saeys2008robust,abeel2010robust,parvandeh2020consensus,frontiersdehp2026,firat2026uveitis}.

We therefore applied frequency-based selection to the same simulated
practitioners. Each retains the genes appearing in the top 50 of at least a
fraction $\pi$ of the 100 model fits they aggregate, and reports that set. Content and provenance are measured as in
Section~\ref{method:provenance}.

One control is essential. Panels selected at a high $\pi$ are short, and short
lists are easier for a model to contain, so provenance can be bought with length
alone. Every stability panel is therefore compared against an averaging panel
\emph{truncated to the same number of genes} (Table~\ref{tab:stability}).

\begin{table}[t]
\centering
\small
\setlength{\tabcolsep}{5pt}
\caption{Selection by recurrence against averaging, at matched panel length.
Each stability panel retains genes appearing in the top 50 of at least $\pi$ of
a practitioner's 100 model fits; each is paired with the averaging panel cut to
the same length, so that containment is not won by brevity. \emph{Provenance} is
the fraction of the reported panel contained in its single best-matching
constituent model. \emph{Content} is recovery of the gene set exclusive to the
lowest-$C$ position, whose size is given in the heading; the intermediate and
high-$C$ positions are omitted because only two and four genes respectively are
exclusive to them under LIME (zero and four under SHAP), which is too few to carry
a percentage. Means over 50 simulated practitioners.}
\label{tab:stability}
\begin{tabular}{@{}lrrr@{\hskip 18pt}rrr@{}}
\toprule
& \multicolumn{3}{c@{\hskip 18pt}}{LIME} & \multicolumn{3}{c}{SHAP} \\
\cmidrule(r{14pt}){2-4}\cmidrule{5-7}
Reported panel & Size & Provenance & Content (11)
               & Size & Provenance & Content (7) \\
\midrule
Averaging, top 50        & 50 & 0.820 & 27\% & 50 & 0.843 & 59\% \\
\addlinespace[3pt]
Stability $\pi = 0.3$    & 56 & 0.786 & \textbf{46\%} & 58 & 0.782 & \textbf{76\%} \\
\quad averaging, matched & 56 & 0.771 & 36\% & 58 & 0.775 & 69\% \\
\addlinespace[3pt]
Stability $\pi = 0.5$    & 37 & 0.947 & 7\%  & 42 & 0.917 & 42\% \\
\quad averaging, matched & 37 & 0.938 & 6\%  & 42 & 0.915 & 41\% \\
\addlinespace[3pt]
Stability $\pi = 0.7$    & 24 & \textbf{1.000} & 0\% & 27 & 0.994 & 11\% \\
\quad averaging, matched & 24 & 1.000 & 0\% & 27 & 0.995 & 6\% \\
\addlinespace[3pt]
Stability $\pi = 0.9$    & 11 & \textbf{1.000} & 0\% & 15 & 1.000 & 0\% \\
\quad averaging, matched & 11 & 1.000 & 0\% & 15 & 1.000 & 0\% \\
\bottomrule
\end{tabular}
\end{table}

\paragraph{The two properties trade off, and no setting delivers both.}
Raising the threshold buys provenance and spends content, monotonically. Under
LIME, provenance rises from $0.786$ at $\pi = 0.3$ to $1.000$ at $\pi = 0.7$
while recovery of the lowest-$C$ exclusive core falls from 46\% to zero; the
panel at $\pi = 0.7$ is twenty-four genes, contained in its entirety by a real
fitted model, and carries none of the content exclusive to the lowest-$C$
position. SHAP reproduces the pattern ($0.782 \to 1.000$ against
$76\% \to 0\%$ between $\pi = 0.3$ and $0.9$). Under LIME at $\pi = 0.3$, a
fifth of the reported panel is absent from the closest fitted model, while more
than half of the lowest-$C$ exclusive content is missing. No tested setting
delivers full provenance and full recovery of that content.

\paragraph{Recurrence does outperform averaging, at low thresholds.}
The comparison at matched length is not uniformly unfavourable to existing
practice and we report where it is not. At $\pi = 0.3$ selection by recurrence
exceeds the equal-length averaging panel on \emph{both} properties: 46\% against
36\% of the lowest-$C$ core at provenance $0.786$ against $0.771$ under LIME,
and 76\% against 69\% at $0.782$ against $0.775$ under SHAP. A practitioner
choosing between the two operations at a fixed panel length is better served by
recurrence than by ranking on mean attribution, and the margin, while modest, is
consistent across both lenses.

\paragraph{At the thresholds in use, the choice of rule ceases to matter.}
That advantage disappears as $\pi$ rises. At $\pi = 0.7$ the two procedures
agree on provenance under both lenses ($1.000$ against $1.000$ under LIME,
$0.994$ against $0.995$ under SHAP) and on content under LIME ($0\%$ against
$0\%$); under SHAP recurrence retains slightly more of the seven-gene lowest-$C$
core ($11\%$ against $6\%$, less than one gene). At $\pi = 0.9$ they agree to
three decimal places. The reason is the grid's own weighting. On the
practitioner's ten-point grid the high-$C$ position holds seven of the ten grid
points and the lowest-$C$ position one, so a gene exclusive to the lowest-$C$
position can never reach a threshold above $0.1$ from that position alone, while
a gene exclusive to the high-$C$ position can reach $\pi = 0.7$ only by appearing
in every high-$C$ fit. Selection at a high threshold is therefore a filter for
breadth across the grid, weighted by the grid itself: it can retain content from
the position the grid samples most heavily---under SHAP, $42\%$ of the four
genes exclusive to the high-$C$ position survive at $\pi = 0.7$---and cannot
retain content from a position it samples lightly. The published
thresholds are high---70\% and 90\% are the conventional choices, and the ones
used in the cross-class comparison above.

What that filter returns here is largely the shared component: the genes every
position attributes. Selection by recurrence at a conventional threshold
therefore arrives at the same destination as averaging by a different route,
at the content the path holds in common rather than the content that
distinguishes its positions.

\paragraph{On the reconstructed population it returns almost nothing.}
Applied to the refits of Section~\ref{sec:lewis}, whose explanations are
dispersed, the same procedure yields a panel of four genes at
$\pi = 0.7$ and, for most simulated practitioners, an empty panel at
$\pi = 0.9$---from an assay of 22{,}835 measured genes. The size-matched
averaging panels are no larger and score no better (provenance $0.960$ and
$1.000$ against $0.994$ and $1.000$). Provenance is undefined for an empty
panel, since it divides by panel size; the $\pi = 0.9$ figures are means over
the minority of practitioners whose panel is non-empty, and the
interpretable result at that threshold is the panel size rather than its
provenance. An almost empty output at this threshold, population and
list size does not by itself show that no stable gene set exists: nothing in the
output distinguishes ``no stable signal exists'' from ``the
threshold was set too high'', and the practitioner receives no indication which
has occurred.

\paragraph{Per-feature reliability is not panel provenance.}
The stability literature scores each feature independently, and does so
effectively: mean per-gene support rises from $0.68$ under averaging to $0.95$ at
$\pi = 0.9$. That is the property these procedures were designed to deliver and
they deliver it. It is a different property from the one a reader of a gene panel
assumes, though the relation between the two depends on the threshold and is
worth stating exactly. At a threshold $\pi$ over $M$ constituent fits, two genes
each appearing in at least $\pi M$ of them must co-occur in at least
$(2\pi - 1)M$; at $\pi = 0.7$ any two selected genes therefore share at least
40\% of the models, and a pairwise failure of joint support is arithmetically
impossible. The failure is available only at panel scale. A panel of $k$ genes,
each absent from at most $(1-\pi)M$ fits, can have those absences cover every
fit once $k \ge 1/(1-\pi)$---four genes at $\pi = 0.7$---so that no single model
contains the whole panel while every gene in it is individually well supported.

Whether that possibility is realised is empirical, and in this population it is
realised at the low thresholds rather than the high ones. At $\pi = 0.7$ some
fitted model does contain the entire twenty-four-gene panel (provenance $1.000$),
whereas at $\pi = 0.3$ a fifth of the reported panel is absent from the closest fit
(provenance $0.786$) although every gene in it recurs in at least 30\% of them.
Per-feature support and panel provenance are therefore distinct quantities that
must be measured separately.
\subsection{When does this occur? Two contrasting populations}
\label{sec:scope}

The result above concerns one pipeline varied along one axis. To bound it, we
applied the same analysis to two further populations on the same data and the
same lens, varying the random seed rather than the regularisation strength.

A deep network (fixed architecture and hyperparameters, seed-only variation,
100 fits per split across 100 splits) produced a single regime under LIME in
all 100 splits. Under SHAP a second cluster was accepted in 79 splits, but it
contained between two and seven of the 100 runs (mean $3.1$), and those runs'
accuracy was indistinguishable from the majority ($0.9864$ against $0.9873$, a
gap of $0.09$ percentage points). Its size makes it a small minority rather than a
second regime of comparable weight; we characterise it below.
The population is concentrated rather than directionless. Its directional
coherence is $\rho = 0.926$ under SHAP (sd $0.026$ across the 100 splits) and
$0.933$ under LIME (sd $0.001$), which by the criterion of
Section~\ref{method:regimes}
places it close to the concentrated end: member attributions agree in direction
and sign. What the most anti-correlated pair of fits (cosine $-0.958$ under SHAP) records is a small
reversed minority, not an absence of structure, and it is the same minority that
SHAP's second cluster accepts. Neither summary statistic separates those two
situations on its own: the mean pairwise cosine of $0.857$ and the coherence
$\rho = 0.926$ satisfy the identity of Section~\ref{method:regimes} to the
precision reported, so they say the same thing. What neither reveals is that the population is a
\emph{mixture}: the same value arises from a homogeneous population in which
every pair sits near $0.86$, and from one in which a large majority is tightly
aligned while a small minority points the other way. What separates those two
cases is the shape of the similarity distribution and the subgroup structure: that pair at $-0.958$, at least
one anti-correlated pair in 95 of the 100 splits, and two internally coherent
parts under the SHAP partition ($\rho = 0.964$ for the majority, $0.865$ for the
minority). What
this pipeline exhibits under seed-only variation is therefore sign-flip
multiplicity---most fits agree on a direction, a few invert it, and no pipeline
parameter identifies in advance which a given fit will be---and not the
dispersed status. The minority is small but internally coherent.
The reversal is observed in SHAP attributions of the class-1 logit; LIME,
which explains the class-1 probability (Section~\ref{method:setup}), resolves a
single region. Adding the same function of the input to both logits leaves every
probability unchanged but can change attributions of the class-1 logit, so a
reversal in these attributions does not by itself establish a reversal in the
classifier's class preference.

A gradient-boosted ensemble under seed-only variation with row and column
subsampling enabled (\texttt{subsample} $= 0.8$, \texttt{colsample\_bytree}
$= 0.8$; 100 fits per split across 100 splits) produced a single
region under LIME in all 100 splits and under SHAP in 99; the single exception
partitions a population whose least similar pair has cosine $0.893$. Mean
pairwise similarity was $0.939$ under LIME and $0.984$ under SHAP, with
split-level means spanning $0.931$--$0.947$ and $0.978$--$0.989$ respectively,
and no pair in any split falling below $0.893$. Accuracy varied across fits
(10--16 distinct values per split), confirming that the fits are genuinely
different models rather than repetitions of one. Directional coherence is
$\rho = 0.970$ under LIME and $0.992$ under SHAP, against $0.933$ and $0.926$
for the deep network and $0.931$ and $0.968$ for the regularisation path. This
population is concentrated on the criterion of Section~\ref{method:regimes}:
admissible variation produces one explanation, and both lenses agree that it
does.

That explanation is not unrelated to the regularisation path. The genes the
boosted ensemble reports reproducibly include most of those specific to the
path's low-$C$ regime and neither of those specific to its high-$C$ regime: at a
70\% stability threshold its panel retains 13 of the 16 low-$C$-specific genes
and 0 of 2 high-$C$-specific ones, and at 90\% it retains 11 of 16, where the
regularisation path's own panel retains none (Table~\ref{tab:crossclass}; LIME,
top 50). The low-$C$ content is therefore not an artefact of a heavy L2 penalty:
a different model class, with a different inductive bias and varied only by
seed, converges on it. The deep network shows this convergence more weakly
(3 of 16 at 70\%). We read this as agreement between model classes, not as
evidence that the low-$C$ content is biologically correct; the regimes'
accuracies remain within the declared tolerance.

Three pipelines on the same data therefore yield three different population
structures: content varying with the regularisation parameter along the
regularisation path, sign-flip multiplicity under seed variation in a deep network,
and a single concentrated explanation under seed variation in a boosted
ensemble. The population structure therefore depends on the model class, the
perturbation scheme and the attribution specification, and is not determined by
this dataset alone; the boosted result is concentration under the seed variation
tested, not a unique explanation across all admissible hyperparameters. The dispersed case---a population with no preferred
direction at all---is exhibited not by any pipeline we constructed but by the
reconstructed biomarker workflow of Section~\ref{sec:lewis}, whose refits have a mean
pairwise cosine of $0.018$.

\subsection{A reconstructed biomarker workflow yields a panel that corresponds to no refit}
\label{sec:lewis}

\begin{figure}[H]
    \centering
    \includegraphics[width=\linewidth]{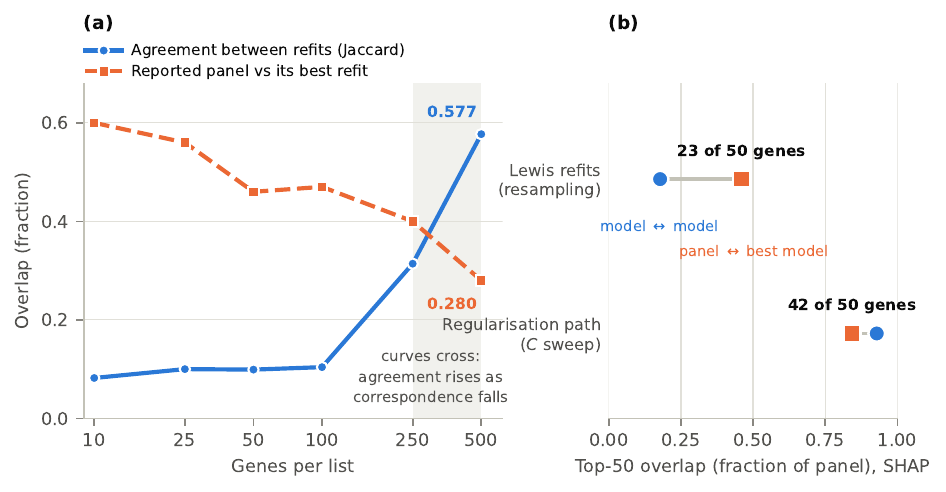}
    \caption{\textbf{Apparent agreement and actual correspondence move in
opposite directions, and the two case studies separate on the same metric.}
\textbf{(a)} Mean pairwise Jaccard overlap between the 20 refits' gene lists
(blue, solid, circles) against the correspondence of the averaged panel to its
single best-matching refit (orange, dashed, squares), as a function of list
length. Agreement rises with length while correspondence falls, and the two
cross between 250 and 500 genes (shaded). Reporting a longer panel therefore
buys the appearance of convergence at the cost of the panel resembling any model
that produced it: over the same range the mean number of refits supporting a
reported gene falls from $7.4$ to $3.5$ of 20.
\textbf{(b)} Top-50 correspondence for both populations, on the same metric and the same definition and under the same lens (SHAP). Squares mark the reported panel's overlap with its best-matching constituent model; circles mark the overlap between two constituent models. On the regularisation path the panel ($0.843$; 42 of 50 genes, and $0.820$, 41 of 50, under LIME) is contained largely in one fitted model, which the practitioner did not choose, while two models of the same regime agree more closely still ($0.928$): the panel is less similar to its best match than such models are to one another. The square is a maximum over fits and the circle an average over pairs. In the audit the panel sits far lower in absolute terms ($0.460$; 23 of 50) while standing well above a near-zero baseline ($0.178$). A wide gap at a low level is what a dispersed population produces: the average sits closer to its best-matching refit than refits typically sit to each other, yet no refit contains more than 23 of its 50 genes.}
    \label{fig:correspondence}
\end{figure}

Sections~\ref{sec:result1}--\ref{sec:stability} concern populations we
generated, under a pipeline we chose. This section audits a reconstruction of a
published biomarker workflow, and compares the two case studies on the same
correspondence metric (Figure~\ref{fig:correspondence}).

\paragraph{Reproduction.}
\citet{lewis2021integration} train a classifier to predict radiation resistance
from gene expression across 915 TCGA tumours and report a ranked gene panel that
is interpreted biologically. We reconstructed their input from public records,
matching all 915 patients and 22{,}835 genes,
 of which 22{,}791 are among the
22{,}818 genes in their own supplementary tables---$99.9\%$ of their reported
set, and $99.8\%$ of ours. We reconstructed their published workflow, including its model class and
hyperparameter-tuning budget, and refit it across 20 independent resamples of the
patients; the 50-gene averaged panel analysed below is built from these refits and
is not the authors' published panel. Following the original workflow, resamples
are stratified by outcome but not grouped by tumour, so the replicate vials of the
three tumours sampled twice (see below) can fall on both sides of a split; this
concerns at most three of the 915 samples.
This varies the data draw rather than a hyperparameter, and therefore probes a
different axis of the admissible pipeline from Section~\ref{sec:result1}---the
axis that any cross-validated or bootstrapped workflow traverses by default.

The 20 refits yield AUROCs of $0.745$--$0.883$ (mean $0.813$). Their
explanations disagree. Mean pairwise Jaccard overlap between
their top-50 gene lists is $0.099$: two refits share about nine genes of fifty.
Signed attribution vectors show almost no directional agreement (mean cosine
$0.018$, median $0.002$, with 49\% of pairs anti-correlated). By the diagnostics
of Section~\ref{method:regimes} this population is dispersed, and unlike the
regularisation path the partition procedure resolves no regimes in it: we find
no coherent alternative explanation for a summary to represent.

\paragraph{The disagreement is between models, not between patients.}
Each refit computes attributions on its own held-out patients, so the observed
disagreement could in principle reflect explaining different patients rather
than different learned mechanisms. Section~\ref{sec:result1} excludes this by
construction; here it must be tested. The 20 test sets are 183-patient draws
from 915 and therefore overlap---median 37 shared patients per pair, range
23--50---which allows the two sources to be separated at a matched number of
patients $m$, without refitting (Table~\ref{tab:lewiscontrol}).

Holding the model fixed and varying the patients leaves the explanation almost
unchanged: two disjoint 40-patient subsets of the same refit's test set yield
top-50 lists overlapping at Jaccard $0.887$. Holding the patients fixed and
varying the model destroys it: two refits scored on the \emph{identical} 40
patients overlap at $0.108$, indistinguishable from the $0.099$ observed when
both model and patients differ. The first figure also establishes that the
measurement is not noise-limited at this sample size, since the within-model Jaccard remains high with 40 patients; the disagreement at $0.108$
is therefore signal. Sample composition does not account for the dispersion.

\begin{table}[H]
\centering
\small
\caption{The dispersion is a property of the models. Mean top-50 Jaccard overlap
under three conditions at a matched number of test patients $m$, over 200 random
draws each. \textbf{A} varies the patients within one refit; \textbf{B} varies
the refit across identical patients; \textbf{C} varies both, the condition
observed in practice. Matching $m$ is necessary because attributions estimated
from fewer patients are noisier, and noise alone depresses agreement; condition
A shows that at these sample sizes it does not. The full-data value is $0.099$.}
\label{tab:lewiscontrol}
\begin{tabular}{@{}lrrrr@{}}
\toprule
$m$ & Pairs & A: same model, & B: different models, & C: different models, \\
    &       & different patients & same patients & different patients \\
\midrule
20 & 190 & 0.856 & 0.097 & 0.103 \\
30 & 178 & 0.869 & 0.092 & 0.101 \\
40 & \phantom{0}60 & 0.887 & 0.108 & 0.099 \\
\bottomrule
\end{tabular}
\end{table}

\paragraph{The reported panel corresponds to no refit.}
The averaged panel follows the convention of the study being audited and of
SHAP summary plots: each refit is summarised by its mean $|$SHAP$|$ per gene over
its own test patients, each such vector is unit-normalised, the 20 are averaged,
and the top 50 are reported; the per-refit lists used for correspondence are the
top 50 of the same per-refit vectors. This is not the rule of the
regularisation-path simulation (Section~\ref{method:phantom}), which averages
signed vectors, and the choice favours the panel rather than the argument: under
signed averaging the panel is more dispersed still, its best-matching refit
containing 15 of its 50 genes rather than 23. Omitting the unit-normalisation
changes the panel by two genes and the three figures below from 23, $7.4$ and 14
to 24, $7.3$ and 15 (Section~\ref{sec:implementation-appendix}).

Applying the correspondence measure of Section~\ref{sec:result3} unchanged, so
that the two case studies are directly comparable: of the 50 genes in the
averaged panel a standard workflow would report, the single most similar refit
contains \textbf{23}. On the regularisation path the corresponding figure is 42
of 50 under SHAP, the lens used here (41 under LIME). More than half of the averaged panel is absent from the closest model
that produced it. Its genes appear in a mean of $7.4$ of the 20 refits, 14 of
the 50 appear in five or fewer, and \emph{no} gene in the panel appears in all
20.

This is the sense in which the panel is a construction of the averaging rather
than a finding of any model, and it differs in degree from what the
regularisation path exhibits: there, the reported panel closely matches a real
model that the practitioner did not choose, sharing 41 or 42 of its 50 genes
with that single fit, and only the remainder is composed; here, no fit comes close, and
there is no model to have chosen.

\paragraph{Why the ratio to the baseline misleads in a dispersed population.}
The panel's correspondence to its closest refit exceeds the correspondence
between two refits ($0.460$ against $0.178$) by a factor of $2.6$, whereas on
the regularisation path the same ratio is $0.91$ under SHAP, the lens used here,
and $1.05$ under LIME. Read naively, the audit
panel would appear better supported than the one from our own simulation. It is
not. The absolute figure carries the information.

\paragraph{Enrichment, against a correct background.}
The assay measures 22{,}835 genes and we supply them as an explicit background,
which is the correct comparison set. Unlike the regularisation path, where the
background is a 1{,}000-gene pre-filtered space and the test has little headroom
(Section~\ref{sec:result2}), this test has the whole assay to work against.
Against it, 10 of the 20 refits yield no significant enrichment at all, the
averaged panel yields none, and the most recurrent term appears in 2 of 20
refits. This is a calibrated enrichment test with ample headroom, and it is null.

\paragraph{Apparent agreement at larger list sizes tracks list volume.}
Agreement between refits does rise if the reported list is made long enough: at
500 genes per model, pairwise Jaccard reaches $0.577$ and the most recurrent enriched
term recurs in all 20 refits (Table~\ref{tab:listsize}). But the terms recurring
at that length are broad housekeeping sets,
and over the same range the averaged panel's correspondence to its best
constituent \emph{falls}, from $0.600$ at ten genes to $0.280$ at five hundred,
while the mean number of refits supporting a reported gene falls from $7.4$ to
$3.5$. The two curves cross (Figure~\ref{fig:correspondence}a). Rising overlap
with falling correspondence is consistent with overlap by volume: longer lists
intersect more because they are longer, while the panel they are compared
against becomes less like any of them.

\paragraph{The finding is not an artefact of the reconstruction.}
Our reproduction reaches a mean AUROC of $0.813$ against the $0.85$ reported by
the original authors; the gap may reflect preprocessing details not recoverable
from the public record. To check that the absence of agreement does not follow
from how we rebuilt the cohort, we refitted the workflow on a second,
independently processed version of the same data, the UCSC Xena TOIL release,
matching 885 of the 912 distinct tumour barcodes (the 915 listed samples
collapse to 912 once replicate vials of the same tumour are removed), with
28{,}613 genes after filtering. Its 20 refits reach AUROCs of $0.733$--$0.903$
(mean $0.798$) and reproduce the result: mean pairwise cosine between signed
attribution vectors $-0.013$ (median $-0.007$, with 52\% of pairs
anti-correlated), mean top-50 Jaccard overlap $0.097$, and panel genes
appearing in a mean of $6.6$ of the 20 refits. The result is therefore a
property of the pipeline rather than of either reconstruction.

\paragraph{Two distinct failures.}
On the regularisation path the reported panel is largely contained in a
real model (41 of 50 genes under LIME, 42 under SHAP), but in one the practitioner did not select, and whose position on the
path is set by the width and centre of their grid. Here no sampled refit
largely contains the panel: more than half its genes are absent from the most
similar refit, and none is common to all twenty. A summary can fail its
population by representing an unchosen member of it or by representing none of
the sampled members, and the reported artefact signals neither.

\subsection{Alternative explanations}
\label{sec:alternatives}

Several objections would, if they held, account for the results above without
explanation multiplicity. Each was tested; this section states each objection
with the measurement that excludes it, and
Appendix~\ref{app:alternatives} gives every test in full, together with the
boundaries of the claim where a test was inconclusive or not performed.

\paragraph{What is excluded.}
\emph{Sample composition} cannot produce the difference between regions, because
every fit within a split is trained on the same patients, so composition cannot
vary between fits that differ only in $C$; the direct test within tissue
sites agrees for ten of the twelve low-$C$-specific genes, which
separate tumour from normal in the same direction in a majority of sites
(Table~\ref{tab:tissue}). \emph{Solver noise} cannot
produce it: the L2-penalised objective is strictly convex, so each value of $C$
has a unique optimum, and every fit converged within its iteration budget (at
most 70 of 1{,}000 lbfgs iterations across all $10^4$
fits; Table~\ref{tab:solver_stability}). \emph{Chance clustering} cannot: under
permuted labels the partition procedure still returns $k^\star > 1$ (mean
$2.00$ over 20 null splits), so a cluster count is not evidence on its
own---what separates the real population from the null is the similarity
distribution, in which every pairwise cosine exceeds $0.5$ against
$51\%$ and $67\%$ of pairs in the two null arms. Nor is the
result an \emph{artefact of one attribution method}: SHAP and LIME place the
boundary within $0.25$ log units of $C$ of each other and agree on run-level
assignment for $72.1\%$ of runs. And \emph{standard metrics do not capture
it}: prediction agreement between regions is $98.97\%$, and clustering models by
their predicted probabilities returns $k^\star = 1$ in 97 of 100 splits, with an
adjusted Rand index of $0.011$ against the attribution partition. Accuracy is not
wholly uninformative about position: under LIME, a fit's accuracy correlates
with the cosine of its attribution to the population's mean direction (Spearman
$0.60$ within splits, against $0.09$ under SHAP), because the low-$C$ regime is
marginally the more accurate. But the whole of that information spans $0.21$
percentage points. The regions are objects in explanation space, and a
monitoring regime that tracks predictive performance would see differences of a
fifth of a percentage point and could not recover the regions from the
predictions.

\paragraph{What is not excluded.}
Two boundaries belong in the main text rather than an appendix.
\emph{Redundancy among correlated features} is not ruled out as the mechanism: a
synthetic control in which one informative signal is duplicated confirms that
redundancy alone generates multiplicity, and compressing the real feature space
to principal components, which removes the correlation between genes, collapses
the population to a single region (Appendix~\ref{app:alternatives}). Compression
also discards most directions of the data and changes the space the explanation
lives in, so this is consistent with, not proof of, correlation as the
mechanism, and we therefore do not claim the
regions correspond to distinct biological mechanisms---only that they carry
different reported content. \emph{Ontology terms} in the regularisation-path
analyses have little headroom against the pre-filtered feature space, which is
itself enriched (Section~\ref{sec:result2}); we therefore report genes, not
terms, there. The retrospective audit of Section~\ref{sec:lewis} is the
exception, and its enrichment result is a calibrated null. Robustness to the
two arbitrary choices in the analysis---the span of the simulated grid and the
positional boundaries---is reported in the appendix; neither moves the result.

\paragraph{Accuracy equivalence follows the convention of the multiplicity literature.}
We treat two regions as predictively interchangeable when their mean held-out
accuracies differ by less than one percentage point, the tolerance used for the
headline results of \citet{marx2020predictive}. Every difference reported here
lies well inside it. Across the full path the largest difference between any two
regions is $0.29$ percentage points under SHAP and $0.21$ under LIME, and close
to the transition it shrinks to $0.055$ points on $C \in [0.04, 0.16]$, about a
quarter of the within-split standard deviation and an eighteenth of the
tolerance. The differences are nonetheless consistent rather than noise: the
low-$C$ region is the more accurate in 96 of 100 splits over the full path and
in 56 of the 69 splits with fits on both sides of the transition, so we do not
claim the regions are statistically indistinguishable, only that they differ by
far less than the margin at which the literature treats models as equivalent.
No claim made here depends on the difference. A reader who holds a stricter
tolerance should read the low-$C$ region as marginally the more accurate
throughout; the explanations still differ, and the more accurate region is not
the one a default grid weights most heavily, since on the full path the low-$C$
region holds 32 of every 100 fits against the high-$C$ region's 68.
\section{Discussion}

The models examined here reach accuracies within a declared tolerance and
differ in their attributions. That much has been established before, in work on
Rashomon sets, predictive multiplicity and underspecification, and in work
treating variable importance as ranging over a class of good models rather than
belonging to one \citep{fisher2019all,donnelly2023rashomon}. This paper takes
three further steps. It shows that the disagreement can have a coordinate: along
the regularisation path, explanatory content changes reproducibly with a single
pipeline parameter, with patients held fixed. It shows that the conventional
summary does not report that population but one position on it, and that the
position is chosen by where the hyperparameter grid sits. And it gives a measure,
model-relative provenance, that makes this visible from fits a practitioner has
already run.

\paragraph{The grid is an undisclosed analytical choice.}
On the regularisation path the reported panel is largely contained in one fitted
model, 41 of its 50 genes under LIME and 42 under SHAP, and which model that is
tracks the centre of the grid, with a rank correlation of $0.92$ across 350
simulated practitioners (Section~\ref{sec:result3}). What the panel does not
share with that model is composed by the averaging: for 43 of 50 practitioners
it combines the two ends' content more fully than any model they fitted, so the
reported explanation belongs to no model whose accuracy was established. The
choice reaches the bench. Two grids centred three decades apart, at accuracies
$0.06$ percentage points apart, give ten-gene shortlists that share $2.4$
candidates once known biology is set aside. Nor is this confined to unusual
settings: the change in content falls inside the range practitioners search, and
\texttt{scikit-learn}'s default grid spans it in every split
(Section~\ref{sec:result1}). A grid is ordinarily reported, where it is reported
at all, as a tuning detail. In a population of this shape it is the parameter
that selects which of several reproducible explanations reaches print, and it is
not reported as such anywhere in the practice surveyed in
Section~\ref{sec:related}.

\paragraph{Why the content is lost, and how it returns.}
The loss has one cause in this pipeline, and the cause bears on the remedy.
Content localised on the parameter's range is removed because the grid weights
its position lightly, and it returns when the grid moves onto it: of the eleven
genes exclusive to the lowest-$C$ position, a grid centred there recovers about
9 and grids centred at $C = 0.3$ or above about 2 to 3, while the four exclusive
to the high-$C$ position move from none to all as the grid moves the other way
(Table~\ref{tab:gridcentre}). \textit{HMGCS2}, among the top 50 genes of the
low-$C$ end in 73 of 100 splits, appears in none of 50 reports from a
conventional grid. The loss is therefore a property of the sweep rather than of
the content, and it is recoverable; a practitioner cannot currently tell that it
has occurred, because it presents in the reported output as absence. Widening
the pool does not supply what the sweep omits: pooling across model classes
lowered the selection frequency of the genes that distinguish positions, and the
pooled panels retained less of that content than the best single population
(Section~\ref{sec:result3}).

\paragraph{A summary can fail its population in two ways.}
The reconstructed biomarker workflow fails differently. Its refits share no
preferred direction (mean pairwise cosine $0.018$), and the averaged panel is
contained in no refit beyond 23 of its 50 genes (Section~\ref{sec:lewis}). On
the regularisation path the panel represents a real member of the population
that the practitioner did not choose; here it represents none of the sampled
members. The two cases call for different readings of the same measure. Relative
to the agreement between models, the dispersed panel looks better supported---a
ratio of $2.6$ against $0.91$ under the same lens---which is why provenance must
be read in absolute terms. Both failures leave the reported artefact unchanged:
a ranked gene list, accurate models and, here, a null enrichment result that a
reader would not connect to either.

\paragraph{Aggregation is not the error; undisclosed aggregation is.}
Averaging fragile signal into a robust summary is legitimate, and on a
homogeneous population it is the right operation. The failure is that nothing in
the reported output indicates which kind of population was summarised. A
practitioner who averages over a homogeneous population and one who averages over
a population whose content changes along a pipeline parameter produce artefacts
that look identical, and the second reader cannot know that an alternative
account existed and was discarded.

\paragraph{Where this matters.}
Where only the prediction is consumed, the explanatory differences studied here
matter less, although accuracy alone does not establish calibration, subgroup
behaviour or robustness under shift. Where the explanation is the deliverable,
the cost is concrete: a published gene panel determines which genes receive
experimental follow-up, which hypotheses are formed and which measurements the
next study includes, and a gene absent from it leaves no trace in the report. The
setting also changes the remedy. Where the mechanism is well understood,
multiplicity can be reduced by constraint, pruning correlated features or
imposing priors that encode knowledge not in question. In discovery the same
operation risks deciding the answer in advance: dropping one of two correlated
genes determines which enters the literature, and filtering by annotation biases
toward mechanisms already annotated. There the appropriate response is to report
the multiplicity rather than remove it.

\paragraph{What the pipeline determines, and what it does not.}
Three pipelines on the same data produce three population structures: content
varying along the regularisation path, a concentrated population with a
sign-reversed minority in a seed-varied deep network, and a concentrated
population in a seed-varied boosted ensemble (Section~\ref{sec:scope}). The
structure depends on the model class, the perturbation scheme and the
attribution specification, not on the dataset alone, so no general claim about
how often the phenomenon occurs can be made from these results, and we make none.
The content is not specific to one model class, however: the boosted ensemble's
single explanation recovers 13 of the 16 genes specific to the low-$C$ end of the
path. How many regions a lens draws along the path is a property of the lens
(Section~\ref{sec:result2}); the aggregation results do not depend on a partition.
Recurrence across splits establishes that the pipeline reliably produces a gene,
not that it reflects the biology that generated the data
\citep{ma2026reconstruction}, so we describe recurrent genes as reproducible
model interpretations and report external corroboration separately.

\paragraph{Limitations.}
The mechanism producing the multiplicity is not established. Redundancy among
correlated features produces it in a synthetic control, and compressing the real
data to principal components removes it, but compression also discards most
directions of the data and changes the space in which the explanation is
expressed. Attributing the effect to correlation would require removing the
correlation among genes while keeping the genes, the penalty and the dimension
unchanged (Section~\ref{sec:alternatives}).

The regularisation path uses an L2 penalty, under which every fit retains all
1{,}000 genes; whether the same dependence holds under a sparsity-inducing
penalty, where genes enter and leave the model along the path, is not tested.
The positions are defined by approximate thresholds in $C$, and the content
exclusive to them is small---11 genes at lowest $C$, four at high $C$ and two in
between---so recovery figures for the high-$C$ and intermediate positions rest
on few genes and are reported with their counts. The regularisation-path
enrichment has little power against the 1{,}000-gene feature space, so the
absence of pathway-level differences between the ends is not evidence that none
exists; it is why we report genes. We do not establish why the lowest-$C$
position carries the most reproducible content, only that it does and that a
conventional sweep weights it least.

The results concern sweeping and aggregating across a grid. We have not measured
how often a realistic model-selection procedure, such as nested
cross-validation or retraining under data drift, lands a deployed model at one
position rather than another; that question decides whether the phenomenon
affects deployed systems as well as published findings, and it is the natural
next experiment. Finally, the composition result of Section~\ref{sec:result3} is
measured against the region signatures and each practitioner's own constituent
fits; the combination it reports exists elsewhere in the fitted population, and
whether an aggregate can support an individual gene or term that no fitted model
supports has not been tested.

\paragraph{Implication for reporting.}
The recommendation is narrow and cheap. Before summarising the explanations a
pipeline produces, establish whether the population supports a single summary:
report its similarity structure, whether its content varies with a pipeline
parameter, what the summary retains relative to its members, and the summary's
provenance---how much of it any single constituent fit contains, and where on the
parameter's range that fit lies. Where a grid was swept and averaged over, report
the grid, because it determines which position the summary represents, and report
the discarded alternatives alongside it. The check costs one analysis over models
already fitted; omitting it leaves a reported explanation whose content was fixed
by a choice no reader can see.
\section*{Acknowledgements}
The EvoXplain framework is the subject of filed UK patent applications. The experiments reported here were run on the University of Hertfordshire High Performance Computing Cluster, whose support is gratefully acknowledged.

\section*{Competing Interests}
EvoXplain is the subject of filed UK patent applications. The author declares no other competing interests.

\bibliographystyle{plainnat}
\bibliography{references}

\appendix
\section{Supplementary tables}

\subsection{Silhouette profiles}
\label{app:silhouette}

Table~\ref{tab:silhouette} gives the full silhouette profile over
$k \in \{2,\dots,8\}$ for both attribution lenses, computed on the
unit-normalised explanation vectors of each split's 100 fits and averaged over
all 100 splits. LIME returns an interior maximum at $k = 2$ in every split,
declining thereafter. SHAP's mean profile peaks at $k = 3$ but is nearly flat
from $k = 3$ to $k = 5$, and the maximising $k$ varies across splits; by the
criterion of Section~\ref{method:regimes} this is a connected structure that
$k$-means segments at a granularity the data do not fix, which is why no result
in this paper is computed on the SHAP partition.

\begin{table}[htbp]
\centering
\small
\caption{Silhouette profile by lens, mean (sd) over 100 splits of 100 fits each.
Bold marks the maximum of the mean profile. LIME attains its maximum at $k=2$ in
100 of 100 splits. SHAP attains it at $k=3$ in 50 of 100, at $k=2$ in 18, at
$k=4$ in 17 and at $k \ge 5$ in 15. The profile is computed without the guards
of Section~\ref{method:regimes}; with them, SHAP's $k^\star$ is 2, 3, 4, 5 and 6
in 19, 56, 18, 4 and 3 splits.}
\label{tab:silhouette}
\footnotesize
\setlength{\tabcolsep}{3pt}
\begin{tabular}{@{}lccccccc@{}}
\toprule
Lens & $k=2$ & $k=3$ & $k=4$ & $k=5$ & $k=6$ & $k=7$ & $k=8$ \\
\midrule
SHAP & 0.641 (0.020) & \textbf{0.665} (0.016) & 0.662 (0.016) & 0.651 (0.016)
     & 0.644 (0.017) & 0.638 (0.019) & 0.631 (0.023) \\
LIME & \textbf{0.252} (0.035) & 0.170 (0.016) & 0.115 (0.015) & 0.092 (0.014)
     & 0.074 (0.017) & 0.083 (0.019) & 0.072 (0.015) \\
\bottomrule
\end{tabular}
\end{table}

\subsection{Tissue-composition audit}
\label{app:tissue}

The classification task pools 9{,}186 primary tumour samples against 727
adjacent-normal samples, and the normal samples are unevenly distributed across
tissue source sites: the ten largest contribute 73, 38, 29, 28, 27, 23, 20, 18,
17 and 17 samples respectively, spanning breast, head and neck, thyroid, liver,
kidney, prostate, colon and lung. Thirty sites have at least eight samples in
each class.

Table~\ref{tab:tissue} reports, for each of the 24 regime-specific genes of
Figure~\ref{fig:regimes}, the number of those 30 sites in which the gene
separates tumour from normal with $|d| > 0.5$ in each direction. A gene whose
apparent effect arose from the over-representation of particular normal tissues
would separate only in those sites. The low-$C$-specific genes mostly separate
in one direction across a majority of sites; several high-$C$-specific genes
separate in opposite directions in different sites, as tissue-restricted genes
would.

\begin{table}[htbp]
\centering
\small
\caption{Within-site tumour-versus-normal separation for the regime-specific
genes of Figure~\ref{fig:regimes}, in the same order. ``Up'' and ``down'' count
sites (of 30) with Cohen's $d > 0.5$ and $d < -0.5$ respectively, computed within
each tissue source site so that composition is held fixed. Strongest three sites
are given with signed $d$; site codes are TCGA tissue source sites.}
\label{tab:tissue}
\begin{tabular}{@{}llccl@{}}
\toprule
Gene & Regime & Up & Down & Strongest sites \\
\midrule
\textit{PLAC9P1} & low-$C$ & 0 & 22 & AA $-3.3$, A6 $-3.2$, HU $-2.4$ \\
\textit{HMGCS2} & low-$C$ & 2 & 24 & BQ $-3.0$, CV $-2.7$, W5 $-2.7$ \\
\textit{CLDN8} & low-$C$ & 6 & 15 & CZ $-8.1$, CJ $-4.6$, BQ $-4.3$ \\
\textit{AADAC} & low-$C$ & 2 & 15 & A7 $-2.4$, E9 $-2.0$, EL $-1.9$ \\
\textit{TCEAL2} & low-$C$ & 0 & 29 & CJ $-2.8$, HU $-2.8$, CZ $-2.8$ \\
\textit{TCEAL5} & low-$C$ & 0 & 25 & CZ $-3.2$, AA $-2.8$, B0 $-2.4$ \\
\textit{GGT6} & low-$C$ & 5 & 20 & CZ $-5.6$, CJ $-4.4$, B0 $-4.0$ \\
\texttt{ENSG00000267328} & low-$C$ & 0 & 23 & A7 $-2.7$, HU $-2.7$, E9 $-2.5$ \\
\textit{HOXC11} & low-$C$ & 21 & 1 & KN $-4.4$, L5 $+3.8$, 22 $+3.4$ \\
\textit{PRAME} & low-$C$ & 20 & 2 & 50 $+3.3$, 22 $+3.3$, 56 $+3.1$ \\
\textit{IGFL2-AS1} & low-$C$ & 18 & 0 & 43 $+3.3$, A6 $+3.3$, 56 $+2.5$ \\
\texttt{ENSG00000144785} & low-$C$ & 17 & 1 & W5 $+2.0$, CW $+1.4$, L5 $+1.4$ \\
\midrule
\textit{CCK} & high-$C$ & 1 & 22 & KN $-2.2$, BC $-1.8$, 50 $-1.7$ \\
\textit{CCDC198} & high-$C$ & 10 & 8 & A6 $+2.0$, AA $+1.6$, 44 $+1.3$ \\
\textit{LINC00992} & high-$C$ & 7 & 12 & HC $+2.5$, A6 $-1.4$, KN $-1.2$ \\
\textit{CHGA} & high-$C$ & 5 & 20 & A6 $-3.0$, AA $-2.7$, L5 $-2.7$ \\
\textit{PITX1} & high-$C$ & 20 & 3 & 22 $+4.6$, W5 $+3.5$, 56 $+3.4$ \\
\textit{TFF2} & high-$C$ & 9 & 10 & EL $-2.9$, KN $-2.7$, BR $-2.1$ \\
\textit{GBA3} & high-$C$ & 1 & 22 & KN $-4.5$, A6 $-3.1$, AA $-3.0$ \\
\textit{NDUFB4P11} & high-$C$ & 1 & 15 & KN $-3.5$, CZ $-2.4$, CJ $-2.1$ \\
\textit{TINAG} & high-$C$ & 13 & 6 & L5 $+2.6$, W5 $+2.0$, BC $+1.6$ \\
\textit{TMEFF2} & high-$C$ & 2 & 24 & AA $-2.9$, HU $-2.7$, KN $-2.5$ \\
\textit{GP2} & high-$C$ & 7 & 11 & KN $-10.2$, CZ $-7.0$, B0 $-4.4$ \\
\texttt{ENSG00000275216} & high-$C$ & 14 & 4 & BR $+2.5$, A6 $+2.4$, L5 $+2.0$ \\
\bottomrule
\end{tabular}
\end{table}

\begingroup
\small
\setlength{\tabcolsep}{4pt}
\begin{longtable}{@{}>{\raggedright\arraybackslash}p{0.14\linewidth}>{\raggedright\arraybackslash}p{0.09\linewidth}>{\raggedright\arraybackslash}p{0.14\linewidth}>{\raggedright\arraybackslash}p{0.33\linewidth}>{\raggedright\arraybackslash}p{0.23\linewidth}@{}}
\caption{Published tumour-versus-normal direction for the regime-specific genes
of Figure~\ref{fig:regimes} with independent evidence, beside their within-site
direction in this cohort (sites of 30 with $d > 0.5$ / $d < -0.5$;
Table~\ref{tab:tissue}). Evidence is protein level (immunohistochemistry,
western blot, immunofluorescence on tissue arrays) unless marked RNA. No
independent evidence was found for \textit{PLAC9P1}, \textit{TCEAL5},
\textit{CCDC198}, \textit{LINC00992}, \textit{GBA3}, \textit{NDUFB4P11},
\textit{TINAG}, \textit{TMEFF2}, \textit{GP2} or the three unannotated loci, and
none for \textit{CCK} from a matched tumour--normal comparison.}
\label{tab:literature} \\
\toprule
Gene & Regime & Here (up/down) & Published direction & Sources \\
\midrule
\endfirsthead
\toprule
Gene & Regime & Here (up/down) & Published direction & Sources \\
\midrule
\endhead
\textit{HMGCS2} & low-$C$ & 2/24 & Lower, colorectal & \citet{camarero2006hmgcs2,hmgcs2crc2019} \\
\textit{PRAME} & low-$C$ & 20/2 & Higher; normal expression confined to testis and proliferative endometrium & \citet{kaczorowski2022prame} \\
\textit{HOXC11} & low-$C$ & 21/1 & Higher, colon (protein and RNA) & \citet{liu2021hoxc11} \\
\textit{TCEAL2} & low-$C$ & 0/29 & Lower, clear-cell renal (protein and RNA) & \citet{Zhou2020TCEAL2} \\
\textit{GGT6} & low-$C$ & 5/20 & Lower, colorectal; lower in breast (RNA) & \citet{Yang2026GGT6,GelenGungor2020GGT6} \\
\textit{CLDN8} & low-$C$ & 6/15 & Conflicting: higher, colorectal; lower, colorectal (RNA) & \citet{cheng2019cldn8,grone2007claudin} \\
\textit{AADAC} & low-$C$ & 2/15 & Tissue-dependent: lower, oesophageal; higher, gastric (RNA) & \citet{Wang2023AADAC,AbdelTawab2022AADAC} \\
\textit{IGFL2-AS1} & low-$C$ & 18/0 & Conflicting, RNA only: higher, colorectal; lower, oesophageal & \citet{Qin2023IGFL2AS1,Marhamati2026IGFL2AS1} \\
\midrule
\textit{CHGA} & high-$C$ & 5/20 & Lower, prostate vs adjacent benign epithelium & \citet{SionVardy2004CHGA} \\
\textit{TFF2} & high-$C$ & 9/10 & Tissue-dependent: lower, gastric; higher, colorectal & \citet{Jiang2014TFF2,Jung2010TFF2,Yu2015TFF2} \\
\textit{PITX1} & high-$C$ & 20/3 & Tissue-dependent: lower, gastric; higher, clear-cell renal (RNA) & \citet{Chen2008PITX1,Zhang2021PITX1} \\
\bottomrule
\end{longtable}
\endgroup

\subsection{List-size sensitivity in the retrospective audit}
\label{app:listsize}

Table~\ref{tab:listsize} reports how agreement and enrichment in the
retrospective audit depend on the number of genes each model contributes.
Apparent convergence at large list sizes coincides with the averaged panel
corresponding to individual models \emph{less} often, not more, which
distinguishes overlap by volume from shared signal.

\begin{table}[htbp]
\centering
\small
\caption{Retrospective audit as a function of reported list size. Enrichment
uses the 22{,}835 measured genes as an explicit background. ``Panel gene
appearance'' is the mean number of the 20 refits in whose own top-$N$ list a
gene of the averaged panel appears.}
\label{tab:listsize}
\begin{tabular}{@{}ccccc@{}}
\toprule
Top-$N$ & Mean Jaccard & Refits with no & Most recurrent & Panel gene \\
        &              & enriched term  & term           & appearance \\
\midrule
50  & 0.099 & 10/20 & 2/20  & 7.4/20 \\
100 & 0.104 & 5/20  & 4/20  & 6.8/20 \\
200 & 0.234 & 2/20  & 15/20 & 5.3/20 \\
500 & 0.577 & 0/20  & 20/20 & 3.5/20 \\
\bottomrule
\end{tabular}
\end{table}

\subsection{Alternative explanations, in full}
\label{app:alternatives}
Each objection summarised in Section~\ref{sec:alternatives} is stated here
with the measurement that addresses it, the settings it was run under, and
the boundary of the claim where a test was inconclusive or not performed.

\paragraph{Sample composition.}
Genes distinguishing the regimes are tissue-restricted, and TCGA's
adjacent-normal samples are unevenly distributed across cancer types, so a
difference between regions could reflect different normal-tissue contrasts
rather than different tumour biology. This is excluded by construction: every
fit within a split is trained on the same patients, so composition cannot vary
between fits that differ only in $C$. The direct test within
tissue sites agrees for ten of the twelve low-$C$-specific genes
(Table~\ref{tab:tissue}); several high-$C$-specific genes (\textit{GP2}, \textit{TFF2},
\textit{TINAG}) are tissue-restricted, and the argument from the within-split design, which
concerns patient membership, does not depend on them.

\paragraph{Model instability or solver noise.}
The L2-penalised logistic loss is strictly convex for every $C > 0$, so a fixed
configuration on fixed data has a unique optimum and run-to-run variation cannot
arise from the optimiser's starting point. What remains is whether each fit
reached that optimum. Every one of the $10^4$ fits was refitted in the current
environment and compared with the archived fit: the
largest number of lbfgs iterations used at any $C$ is 70 of the 1{,}000 allowed, and no fit raised a convergence warning
(Table~\ref{tab:solver_stability}); every refit reproduced the archived held-out
accuracy exactly and its predicted probabilities to within $5 \times 10^{-7}$.
{} Solver noise therefore does not account for the regime separation.

In addition, all regions fall within the declared accuracy tolerance: the largest difference in mean held-out accuracy between any two regions is $0.0029$ under SHAP and $0.0021$ under LIME---that is, $0.29$ and $0.21$ percentage points, the latter being the full-path difference reported in Section~\ref{sec:result1}. The tolerance is the one percentage point used for the headline results of \citet{marx2020predictive} (Section~\ref{sec:alternatives}). The difference is nonetheless systematic: under LIME the low-$C$ region is the more accurate in 96 of 100 splits (per-split difference $+0.0021$, sd $0.0013$, range $-0.0012$ to $+0.0054$), so the regions are \emph{not} statistically indistinguishable, and we do not claim they are; a reader who holds a stricter tolerance should read the low-$C$ region as marginally the more accurate throughout. That reading does not affect any claim made here---the explanations still differ, and the more accurate region is not the one a default grid weights most heavily, since on the full path the low-$C$ region holds 32 of every 100 fits against the high-$C$ region's 68. For scale, the difference is $1.02$ times the mean spread across runs within a single split, so it is of the order of the run-to-run variation a practitioner already absorbs rather than beneath it. It depends on the window compared: $+0.0021$ over the full path, $+0.0024$ on $C \in [0.01, 10]$ (low-$C$ more accurate in 100 of 100 splits), $+0.0015$ on $[0.016, 0.4]$ (97 of 100), and $+0.00055$ on $[0.04, 0.16]$, where 56 of the 69 splits with fits on both sides of the transition favour the low-$C$ region. Region assignment is monotone in $C$ within a split after ordering regions by median $C$ (weak pass rate $1.000$ under both lenses; the stricter non-overlap criterion passes in $85.7\%$ of splits under SHAP and $98\%$ under LIME). Each split fits one model per $C$ value, so there is no within-$C$ replication to test for stability; strict convexity and the convergence audit (Table~\ref{tab:solver_stability}) make such replication unnecessary.

\paragraph{Label structure or chance clustering.}
The controls in this paragraph were run in a separate configuration from the
main analysis: 20 splits with a 20-point $C$-grid under an $\ell_2$ penalty,
with explanation vectors pooled across splits and clustered as a single set of
400. They bear on whether the phenomenon requires real labels, not on the
partition structure reported in Section~\ref{sec:result2}, and their cluster
counts are not comparable with those reported there.

Under permuted labels the pipeline's held-out accuracy falls to $0.9061$, below
the $0.9310$ obtained by always predicting the majority class, so the models learn
nothing beyond the class balance, and the resulting explanation vectors are
near-orthogonal to those from the real labels (maximum cross-cosine
$0.0791$ over 20 splits). The structure is therefore not an artefact of
the feature matrix or of the $C$-grid alone.

Chance clustering deserves a stronger statement than a pass mark. Under
permuted labels the partition procedure still returns $k^\star > 1$: mean
$k^\star = 2.00$ across the 20 null splits. A value of $k^\star$ greater
than one is therefore not by itself evidence of anything. What separates the
real population from the null is the similarity distribution, not the cluster
count: in the real population every pairwise cosine exceeds $0.5$ (mean
$0.754$, sd $0.067$, minimum $0.569$), whereas in the two
null arms only $51\%$ and $67\%$ of pairs do (mean
$0.511$ and $0.536$), and in a randomised control every pair is
near zero (mean $-0.0002$). The Kolmogorov--Smirnov distance between the
real distribution of pairwise cosines and each null arm's is
$D = 0.948$ and $0.941$; we report the distance as a descriptive
separation and attach no $p$-value to it, because the pairs share models and are
not independent observations.
This is a further reason the aggregation results in
Sections~\ref{sec:result3} and~\ref{sec:stability} are computed without
reference to any partition: a statistic that returns two regions on permuted
labels is not a sound foundation for a claim about what a summary discards.
\paragraph{Attribution artefact.}
Two attribution methods with different formulations resolve the same
population. LIME returns an interior silhouette optimum in every split and SHAP
a near-plateau, SHAP resolving more regions than LIME in 81 of 100 splits; both
place a boundary at the same location within $0.25$ log units of $C$, and
run-level assignment agrees for $72.1\%$ of runs. Cross-lens gene overlap
between matched regions is $0.607$ (Jaccard, top 30 genes). Raw cross-lens
centroid cosine is low ($0.172$); the agreement is in structure, not magnitude.
We do not claim that the two lenses agree on how many regions exist---they do
not, and Section~\ref{sec:result2} treats that disagreement as informative
rather than as error.

\paragraph{Standard metrics already capture this.}
They do not. Between the regions LIME resolves, prediction agreement over
cross-region pairs of fits is $98.97\%$ and their mean absolute difference in
predicted probability is $0.025$, while the difference in mean held-out accuracy
between the extreme regions is $0.21$ percentage points and the top-50 genes of
the lowest- and highest-$C$ profiles overlap at a Jaccard index of $0.505$; under
SHAP the corresponding figures are $99.06\%$, $0.022$, $0.21$ percentage points
and $0.537$. Averaged over region-crossing pairs of individual fits rather than
over regional means, the accuracy difference rises only to $0.32$ points under
LIME and $0.34$ under SHAP. Accuracy is not wholly uninformative about
attribution: the within-split Spearman correlation between a fit's accuracy and
the cosine of its attribution to the population's mean direction is $0.60$
(sd $0.20$ across splits) under LIME and $0.09$ (sd $0.19$) under SHAP. But the
range of accuracy that correlation spans is a fifth of a percentage point, and
restricting to the highest-accuracy quartile of models still leaves an
attribution spread ($1 - $ minimum pairwise cosine) of $0.148$ under LIME and
$0.141$ under SHAP. A monitoring regime that tracks predictive performance will
not observe the regions.

\paragraph{The regions are not visible in prediction space.}
Models in different regions do make measurably different predictions: pairwise
disagreement is $2.7$ times higher between regions than within them under LIME
and $3.3$ times higher under SHAP (a ratio of means over dependent pairs of
models, reported descriptively). But
clustering the models by their predicted probabilities does not recover the
attribution regions: probability-space clustering returns $k^\star = 1$ in 97 of
100 splits, and the adjusted Rand index between probability-space and
attribution-space partitions is $0.011$ under LIME and $0.014$ under SHAP. The
regions are objects in explanation space. This supports a practical claim---the
probability-space clustering we tested does not recover them---and bounds the
theoretical one: we do not claim the models implement different decision
functions in any stronger sense.

\paragraph{Redundancy among correlated features.}
Where two features carry overlapping signal, a penalty may
divide weight between them differently at different strengths, and the resulting explanation multiplicity would be a property of
feature correlation rather than of distinct mechanisms. A synthetic control in
which a single informative signal is duplicated confirms that redundancy alone
generates multiplicity. With two independent features that are both required,
so that no redundant pathway exists, the population is unimodal
($k^\star = 1$). With two interchangeable copies of one latent signal the
partition procedure returns $k^\star = 5$ over 50 runs, and the five regions
differ only in how they divide the attribution between the two copies---$0.98$,
$0.74$, $0.49$, $0.32$ and $0.05$ of the signal mass on the first copy, in
regions of $23$, $9$, $5$, $9$ and $4$ runs---with no measurable attribution on
the eighteen noise features in any of them. The dataset is constructed with two
ground-truth alternatives and the procedure returns five, which is itself the
point: this is a continuum of admissible combinations segmented by the
partition procedure, not a set of discrete alternatives---a caution about
reading $k^\star$ as a count of mechanisms. This control was run on a single
synthetic split and is reported as an illustration rather than as a replicated
result.

We have not established that correlation among the genes is the mechanism, but
perturbations of the real data point towards it. In a separate configuration of
10 splits with a 20-point $C$-grid, ablating the 100 highest-weighted features,
or injecting noise into them, leaves the population much as it was---mean
pairwise cosine $0.926$ and $0.930$ against $0.932$ for the unperturbed baseline,
with the partition procedure still returning $k^\star > 1$ (mean $2.2$ and $2.4$
against $2.6$)---so the multiplicity is not carried by a few features.
Compressing the feature space to 50 or 100 principal components, which removes
the correlation between inputs, collapses it: $k^\star = 1$ in every split and
mean pairwise cosine $0.992$ and $0.993$, at accuracies within $0.4$ percentage
points of the baseline. This is consistent with correlated features being the
precondition for the multiplicity, as in the synthetic control above, but it is
not decisive. The L2 penalty is unchanged by a rotation of the inputs, so it is
not the rotation that matters; compression also discards all but 50 or 100
directions of the data and moves the explanation into a space of 50 or 100
components rather than 1{,}000 genes, and either could make the population more
coherent on its own. A decisive test would remove the correlation among genes
while keeping the genes, the penalty and the dimension unchanged, which we have
not attempted.
The claim in this paper concerns what the pipeline produces and what the
conventional summary retains, and does not depend on the mechanism generating
the multiplicity.

\paragraph{Feature space and enrichment background.}
Two choices outside the model shape what an explanation can contain, and both
are held fixed here rather than varied. The first is the feature space. The
regularisation-path analyses use the 1{,}000 highest-variance genes, selected
from the full transcriptome and standardised per gene before any split is
drawn (Section~\ref{method:setup}), and identical across all
$10^4$ fits and all 100 splits. Explanations are therefore defined relative to that space: a
gene excluded by the pre-filter cannot appear in any region's profile, so its
absence is uninformative, and a different pre-filter would define a different
explanation space in which the number and content of regions could differ.
The multiplicity reported in Section~\ref{sec:result1} is a property of this
pipeline including its pre-filter, and we make no claim about what a different
pre-filter would yield.

The second is the enrichment background, and we tested it rather than
assuming it. Running the regularisation-path enrichment against the 1{,}000-gene
feature space as an explicit background, in place of g:Profiler's default
genome-wide domain, returns one significant term across the 200 regime panels.
Section~\ref{sec:result2} gives the reason: the feature space is
itself significantly enriched for 184 terms, including extracellular and
epithelial terms, so the headroom for further enrichment against it is small
and a 50-gene query rarely clears it, while the metabolic and secretory
terms---not filter-enriched---do not reach significance from a 50-gene query
against a 1{,}000-gene domain; the planted-signal check there confirms that only
large coherent sets are detectable.

We therefore report no ontology term from the regularisation-path analysis as a
finding. The retrospective audit of Section~\ref{sec:lewis} is different: there
the custom background is the measured gene set, the test has ample headroom, and
its result is null. Three things are unaffected by the background choice.
Section~\ref{sec:result1} is computed on attribution vectors and involves no
ontology. The gene-level content of each region---which genes enter which
region's profile, and how far each position's reproducible core is exclusive to
it (Table~\ref{tab:cores})---is measured directly on ranked attributions. And
the within-site checks of Table~\ref{tab:tissue} and the external corroboration
of Section~\ref{sec:result2} are gene-level.
 
One point of interpretation follows from how the exclusive sets are defined,
and we state it because it is easy to over-read. A set exclusive to one position
is exclusive with respect to the \emph{profiles}, which are means over that
position's fits. It is not a claim that the two sets never appear together in
one model, and they do. Judged on its own top-50 list---the same length at
which the cores are defined---an individual fit carries a mean of $32\%$ of the
eleven-gene low-$C$-exclusive set and $45\%$ of the four-gene high-$C$-exclusive
set simultaneously, and the single fit that balances them best carries six of
the eleven and three of the four, in one model. The positions differ in
what they concentrate on, not in what they are capable of containing. This is
also why a reported panel can carry much of both sets: the combination is
available in the population, though, as Section~\ref{sec:result3} shows,
usually not among the fits a practitioner averaged.

\paragraph{The choice of grid span.}
The practitioner simulation uses a ten-point log-spaced grid over
$C \in [0.01, 10]$, which spans the transition, so the measured loss could in
principle be a property of that particular span. Section~\ref{sec:result3}
addresses the grid's \emph{centre} directly and reports it as a result rather
than as a control. Here we record the complementary sweep over spans, measured
at gene level---which, unlike term recurrence, does not depend on the enrichment
background.
 
The loss is not a matter of gene content being discarded wholesale. Against
region signatures defined as genes appearing in a region's profile in at least
70\% of splits (25 shared, 18 low-$C$-specific, 4 high-$C$-specific), the reported
panel recovers the shared component completely and most of both region-specific
sets under every straddling grid: $68\%$ and $88\%$ for the grid used here, and
similar figures for a grid confined to $C \in [0.05, 1]$ or matching
\texttt{scikit-learn}'s \texttt{LogisticRegressionCV} default
(Table~\ref{tab:sigexcl}). The high-$C$-specific figures rest on four genes and
should be read with that count in view. Recovery of the content \emph{exclusive}
to a position is a different matter, and Table~\ref{tab:sigexcl} reports it on
the same panels: $25\%$ for the lowest-$C$ position against $88\%$ for the
high-$C$ one. What the loss consists of instead is set out in
Section~\ref{sec:result3}: content localised on the path is outvoted by the
positions the grid weights more heavily.
 
Grids confined to one end are the more lossy choice, not the less. A grid
over $C \in [0.001, 0.1]$ draws 92\% of its fits from the low-$C$ region and
recovers none of the high-$C$-specific signature; a grid over $C \in [1, 10^{3}]$
samples the high-$C$ region exclusively and recovers $36\%$ of the
low-$C$-specific signature (Table~\ref{tab:sigexcl}).
A practitioner who narrows their grid to avoid the transition---which they could
only do knowingly---would report less of the population, not more. No span
tested, and no centre in Table~\ref{tab:gridcentre}, returns all of both
exclusive sets.
 
\paragraph{The positional boundaries.}
The decomposition used in Sections~\ref{sec:result2} to~\ref{sec:stability}
divides the path at $C \approx 0.016$ and $C \approx 0.081$, and both values are
taken from the partition analyses of Sections~\ref{sec:result1}
and~\ref{sec:result2}. A decomposition that avoids clustering at the point of
measurement but inherits its boundaries from a clustering result is only partly
independent of it, and we tested whether the finding survives boundaries chosen
without reference to any partition. Repeating the analysis with the path divided
at $C = 0.01$ and $C = 0.1$---round values spanning the same range, derived from
nothing---moves a share of the fits between positions: the three
positions then hold 17, 17 and 66 of the 100 fits per split, against 20, 12 and
68 under the estimated boundaries. The result does not move with them. The
centre-level correlation between grid centre and median best-matching model is
$0.960$ under both, and recovery of the content exclusive to the most heavily
regularised position still falls monotonically as the grid moves away from it,
from 85\% to 18\% (86\% to 16\% under the estimated boundaries). The reproducible
core sizes are likewise stable, at 45 and 28 genes for the lowest-$C$ and
high-$C$ positions against 45 and 30.

The agreement between the two correlations is exact rather than approximate, and
this is not a coincidence. The correspondence between a practitioner's grid
centre and the fitted model their panel matches is computed from attribution
vectors and model identities alone; no boundary, partition or threshold enters
it. Only the recovery columns depend on the decomposition, and those are robust
to it.

\paragraph{An argument against collinearity that we do not use.}
In the refit population of Section~\ref{sec:lewis}, genes whose attributed sign
varies across refits carry stronger marginal associations with the outcome than
genes whose sign is stable---mean $|r|$ of $0.32$ against $0.22$ over all
1{,}216 genes ($p = 3 \times 10^{-21}$, Mann--Whitney). This is sometimes
offered as evidence against a collinearity explanation, on the grounds that
instability is confined to weak features. We decline the inference twice over.

First, the comparison is confounded by how often a gene is selected at all. A
gene entering only one refit cannot be observed to flip, and is recorded as
stable by construction; the sign-stable group consists almost entirely of such
genes, appearing in a mean of $1.19$ refits against $4.37$ for the sign-varying
group. Restricting to genes that could have flipped shrinks the difference from
$0.10$ to $0.04$, and within a matched coverage stratum to $0.01$. Among the 94
genes entering five or more of the 20 refits, 93 vary in sign. Second, even were
the direction real, sign reversal in the presence of correlated predictors is
the expected behaviour of a suppressor relationship, and occurs for strong
features as readily as weak ones. The instability is measured; its cause is not
established.

\subsection{Implementation and availability}
\label{sec:implementation-appendix}

\paragraph{Attribution configuration.}
Attributions are computed for a fixed set of 200 held-out patients per split,
chosen once per split with a fixed seed as those whose class-1 probability
under a reference random forest lies in $[0.45, 0.55]$, or the 200 nearest
$0.5$ if the band holds fewer; the same patients are shared by every fit in
that split, and a run's attribution vector is the mean of its per-patient
attributions. SHAP uses the explainer matched to the model class:
\texttt{LinearExplainer} for logistic regression, whose target is the linear
predictor (the log-odds); \texttt{TreeExplainer} for the gradient-boosted
ensemble, whose target is the raw margin; and \texttt{DeepExplainer} for the
network, whose target is the class-1 output of the two-output layer before the
softmax. The SHAP background is 100 training patients drawn once per split.
LIME uses \texttt{LimeTabularExplainer} with 500 perturbations per patient for
the regularisation path and the boosted ensemble and the library default of
1{,}000 for the network, no discretisation, and the training fold as reference
distribution; its local surrogate is fitted to the class-1 predicted
probability in every model class. For logistic regression and the tree
ensemble the two lenses' targets are related by a monotone transform of one
scalar; for the network the class-1 logit and the probability
$\sigma(z_1 - z_0)$ are not, so the attributed sign of a feature may differ
between them. Within a lens, sign agreement, coherence $\rho$ and region
structure are always computed between fits under the same target. Because
per-gene standardisation sets the effective L2 penalty on each gene,
train-only preprocessing could shift absolute accuracies as well as the
regularisation path itself.
\paragraph{Enrichment configuration.}
Enrichment was performed with \texttt{gprofiler-official} 1.0.0 against
g:Profiler database release \texttt{e114\_eg62\_p19\_27110d83}, with \\ \texttt{organism=hsapiens}, \texttt{significance\_threshold\_method=g\_SCS}, \texttt{user\_threshold=0.05} and sources GO:BP, GO:MF, GO:CC, KEGG and REAC.
Term--gene intersections were retained throughout. The regularisation-path
enrichment was run against both g:Profiler's default domain \\
(\texttt{domain\_scope=annotated}) and, as reported, the 1{,}000-gene feature
space as an explicit background (\texttt{domain\_scope=custom}).
For the retrospective audit the 22{,}835 measured genes were supplied as an
explicit background with \texttt{domain\_scope=custom}; using the default genome-wide domain instead
returns terms reflecting assay coverage rather than model content, and changes
the reported figures substantially. Ontology annotations change between
database releases, and recurrence figures are not reproducible without matching
both the client and database versions above.

A g:Profiler request that fails returns no terms, and a script that catches the
failure and continues would record it indistinguishably from a biological null.
The two archived producers are exposed to this differently.
\texttt{tcga\_basin\_recurrence\_check.py}, which produced the per-split records
under both backgrounds, does not catch request errors: a failed request aborts
the run before the record file is written, so a complete file establishes that
every request in it returned. \texttt{enrich\_phantoms.py}, which produced the
per-practitioner enrichment records, originally caught
request errors and wrote an empty result; the archived version records a
per-request status and excludes failures from every denominator. For the
records used here, \texttt{check\_enrichment\_zeros.py} counts the
zero-term records, the only place a failure could hide: 0 of 50 per-practitioner
records, 0 of 200 per-region requests against the default domain, and 199 of 200
against the feature-space background. The last is the single-term result of
Section~\ref{sec:result2}; it was produced by the script without a
silent-failure path, and is therefore a null of the test, not of the network.

The per-regime tallies are pooled across splits by the index of the
regime centroid within each split, and that index is a $k$-means label, which
carries no guaranteed meaning across independently fitted splits. We therefore
checked it rather than assuming it (\texttt{check\_basin\_ordering.py}): in all
100 splits under LIME the label order is the same, index 0 holding the least
regularised region (median $C$ between $6.58$ and $16.3$) and index 1 the most
regularised (median $C$ between $0.0061$ and $0.0152$), and under SHAP the same
holds for all 56 splits with $k^\star = 3$. Pooling by index therefore pooled
by region in every split, and the figures name the regions correctly. The archived
recurrence script now indexes regimes by median $C$ by default and records
which convention a summary was written under, and the figure script resolves
regimes by that record rather than by position, so a regenerated summary cannot
silently exchange the two regimes.

Two archived scripts build the audit's averaged panel, and an earlier version
of one of them did so without the per-refit unit-normalisation.
\texttt{check\_lewis\_panel\_definitions.py} builds the panel under both rules
from the same 20 refits: they share 48 of 50 genes (\textit{B3GALT1} and
\textit{CAMK2G} under the plain mean in place of \textit{N4BP3} and
\textit{TUBB2B} under the normalised mean), and the correspondence figures are
24 of 50, 7.3 refits per gene and 15 genes in five or fewer under the plain
mean against 23, 7.4 and 14 under the normalised mean reported in
Section~\ref{sec:lewis}. Both scripts now normalise by default; the earlier
rule remains available as \texttt{--normalise 0}. The per-refit norms of mean
$|$SHAP$|$ span a factor of $2.6$, which is why the two rules do not coincide
exactly. Averaging signed SHAP vectors instead, the rule of the
regularisation-path simulation, gives a more dispersed panel still: 15 of 50 in
the best-matching refit, $4.3$ refits per gene and 35 genes in five or fewer.

The aggregation analyses use $N = 50$ throughout and a reproducibility threshold
of $\theta = 0.7$. Each condition simulates 50 practitioners, each drawing 10
data splits and scanning a ten-point log-spaced grid within each. The span sweep
varies the grid endpoints at fixed point count; the centre sweep fixes the grid
width at $\pm 1.5$ decades and slides its centre. Recurrence selection is
reported at $\pi \in \{0.3, 0.5, 0.7, 0.9\}$, each compared against an averaging
panel truncated to the same length. The sample control of
Section~\ref{method:patientcontrol} uses 200 random draws per condition per
matched size $m \in \{20, 30, 40\}$. All simulations use a single fixed random
seed, recorded in the analysis code, and refit no models: every attribution
vector is indexed out of the precomputed population.

Analysis code is available at \url{https://github.com/bensmailchama-boop/evoxplain-reproduction/}. Per-split explanation vectors,
region assignments and refit outputs are archived at DOI \url{10.5281/zenodo.22769609}.
Table~\ref{tab:provenance} maps each figure and table in this paper, and each
control reported in Section~\ref{sec:alternatives}, to the script and the
archived output directory that produced it. Every quantity quoted in the text
is produced by one of these scripts; where a figure in an earlier draft could
not be reproduced from the archived outputs, the figure has been corrected to
the reproducible value rather than the reverse.

\begingroup
\small
\begin{longtable}{@{}>{\raggedright\arraybackslash}p{0.20\linewidth}>{\raggedright\arraybackslash}p{0.36\linewidth}>{\raggedright\arraybackslash}p{0.36\linewidth}@{}}
\caption{Source of each reported result. Each reported figure, table and
control, the script that produced it, and the archived output it was read from.
Paths are relative to the archive's results root. Every row that does not name
another source reads the converged regularisation-path run,
\texttt{rerun\_sep2026/tcga\_lr\_cgrid\_shap\_lime\_100x100} (100 splits
$\times$ 100 fits, both lenses), and writes its log to
\texttt{rerun\_sep2026/regen/logs/}; the log name is given.}
\label{tab:provenance} \\
\toprule
Reported item & Script & Archived output \\
\midrule
\endfirsthead
\toprule
Reported item & Script & Archived output \\
\midrule
\endhead
\multicolumn{3}{@{}l}{\textit{Figures}} \\
Fig.~\ref{fig:population} & \path{make_figures.py} (\path{figure1}), via \path{make_gene_figures.py} & \path{figures.log} \\
Fig.~\ref{fig:regimes} & \path{build_gene_recurrence.py}, \path{make_gene_figures.py} & \path{rerun_sep2026/recurrence_lime}; \path{regen/gene_recurrence.json} \\
Fig.~\ref{fig:aggregation} & \path{simulate_practitioner_phantoms.py} (two grids), \path{build_gene_recurrence.py}, \path{make_fig3_two_grids.py} & \path{rerun_sep2026/phantoms_lime_c0p01}, \path{phantoms_lime_c10} \\
Fig.~\ref{fig:correspondence} & \path{fig_lewis_correspondence.py} & \path{lewis_audit_faithful/_individual}; path values (SHAP) from \path{panel_prov_shap.log} \\
\midrule
\multicolumn{3}{@{}l}{\textit{Tables}} \\
\ref{tab:partitions} & \path{path_structure_check.py}, \path{shap_k3_composition.py} & \path{partitions_lime.log}, \path{partitions_shap.log}, \path{shap_k3_comp_100.log} \\
\ref{tab:cores} & \path{grid_centre_sweep.py} & \path{gridcentre_lime.log}, \path{gridcentre_shap.log} \\
\ref{tab:gridcentre} & \path{grid_centre_sweep.py} & \path{gridcentre_lime.log}, \path{gridcentre_shap.log} \\
\ref{tab:sigexcl} & \path{check_signature_vs_exclusive.py} & \path{sigexcl.log} \\
\ref{tab:stability} & \path{stability_provenance.py} & \path{stability_lime.log}, \path{stability_shap.log} \\
\ref{tab:lewiscontrol} & \path{lewis_patient_control.py} & \path{lewis_audit_faithful/_individual} \\
\ref{tab:silhouette} & \path{sil_profile.py}; $k^\star$ counts by \path{check_basin_ordering.py} & \path{sil_shap.log}, \path{sil_lime.log}, \path{basin_order_shap.log} \\
\ref{tab:tissue} & \path{tissue_composition_check_v2.py} & \path{tissue_v2.log} \\
\ref{tab:listsize} & \path{lewis_listsize_sweep.py} & \path{lewis_audit_faithful/_individual} \\
\ref{tab:crossclass} & \path{cross_class_consensus.py}; population check by \path{xgb_control_check.py} & \path{cross_class.log}; DNN and XGB aggregates from \path{final_freeze}, which the rerun did not affect \\
\ref{tab:solver_stability} & \path{convergence_audit.py} (\path{--run} and \path{--out} set to the rerun) & \path{rerun_sep2026/tcga_lr_cgrid_convergence} (one CSV per split); \path{convergence_summary.txt} \\
\midrule
\multicolumn{3}{@{}l}{\textit{Results in the text}} \\
Regime similarity, accuracy, transition & \path{check_result1_gaps.py}, \path{regime_accuracy.py}, \path{path_structure_check.py} & \path{result1_gaps.log}, \path{regime_accuracy.log}, \path{partitions_lime.log} \\
Directional coherence & \path{check_rho_accgap.py} & \path{rho_lime.log}, \path{rho_shap.log} \\
Default-grid points by regime; low-$C$ share of each grid & \path{check_grid_regimes.py} & \path{grid_regimes.log} \\
Panel provenance, landing point, joint coverage & \path{panel_provenance.py} (\path{--lens lime}, \path{--lens shap}) & \path{panel_prov.log}, \path{panel_prov_shap.log} \\
Progress along the path; transition width and midpoint & \path{check_path_transition.py} & \path{rerun_sep2026/transition_check/} \newline \path{path_transition_lime.log}, \path{path_transition_shap.log} \\
Aggregation rules, participation ratio & \path{normalisation_bias.py} & \path{normalisation.log}, \path{normalisation_shap.log} \\
Follow-up prioritisation & \path{check_prioritisation.py} & \path{prioritisation*.log} (four runs) \\
Feature-space enrichment & \path{check_featurespace_enrichment.py} & \path{featurespace.log}, \path{recur_featurebg.log} \\
Enrichment power; competitive gene-set test & \path{check_featurebg_power.py}, \path{check_regime_pathway_shift.py} & \path{featurebg_power.log}; \path{pathway_shift.log}, \path{regime_pathway_shift.json}, \path{gene_sets_featurespace.json} \\
\midrule
\multicolumn{3}{@{}l}{\textit{Controls in Section~\ref{sec:alternatives}}} \\
Sample composition & \path{tissue_composition_check_v2.py} & \path{tissue_v2.log} \\
Region accuracy gaps; monotonicity & \path{tcga_test3_model_instability.py} & \path{rerun_sep2026/falsification/} \newline \path{tcga_test3_results.json} \\
Permuted labels; null $k^\star$; null arms; randomised control; KS tests & \path{tcga_test2_null_noise.py} & \path{rerun_sep2026/falsification/} \newline \path{tcga_test2_results.json}, \path{test2.out} \\
Cross-lens agreement & \path{tcga_test12_attribution_lenses.py} & \path{rerun_sep2026/falsification/} \newline \path{tcga_test12_results.json} \\
Prediction-space structure; ARI & \path{tcga_test5_ranking_functional.py} & \path{rerun_sep2026/falsification/} \newline \path{tcga_test5_results.json} \\
Ablation, noise, PCA compression & \path{tcga_test7_data_degeneracy.py} & \path{rerun_sep2026/falsification/} \newline \path{tcga_test7_results.json}, \path{test7.out} \\
Redundancy (synthetic) & \path{evoxplain_core_engine.py} \newline (\path{synth_k1}, \path{synth_k2}) & \path{synth_validation/synth_k1_matched}, \path{synth_k2_matched} \\
Standard metrics & \path{check_standard_metrics.py} & \path{std_metrics_lime.log}, \path{std_metrics_shap.log} \\
Enrichment request completeness (0/50, 0/200, 199/200) & \path{check_enrichment_zeros.py} & \path{zeros_splits.log}, \path{zeros_phantoms.log} \\
Regime index consistency (100/100 LIME, 56/56 SHAP) & \path{check_basin_ordering.py} & \path{basin_order_lime.log}, \path{basin_order_shap.log} \\
Exclusive-core co-occurrence & \path{check_core_cooccurrence.py} & \path{cooccurrence.log} \\
Grid span sweep & \path{grid_span_sweep.py} & \path{gridspan.log} \\
Round positional boundaries & \path{grid_centre_sweep.py} (\path{--ultra 0.01 --trans 0.1}) & \path{gridcentre_round.log} \\
Sign variation and marginal association & \path{evoxplain_suppressor_diagnostic.py}, \path{check_suppressor.py} & \path{lewis_audit_faithful/provenance_audit/} \newline \path{suppressor_diagnostic.csv} \\
\midrule
\multicolumn{3}{@{}l}{\textit{Retrospective audit (Section~\ref{sec:lewis})}} \\
Cohort reconstruction & \path{match_lewis_samples.py} & \path{data/lewis_audit} \\
Second reconstruction (Xena) & \path{lewis_build_genelists.py} on \path{lewis_audit_pilot/_individual} & \path{outputs/lewis_xena/xena_agreement.log} \\
Gene lists and agreement statistics & \path{lewis_build_genelists.py} & \path{lewis_audit_faithful/_individual} \\
Gene-set overlap & \path{check_lewis_gene_overlap.py} & \path{lewis_background.json} \\
Refit population & \path{lewis_baselearner_faithful.py} & \path{lewis_audit_faithful/_individual} \\
Panel correspondence & \path{lewis_provenance.py}, \path{evoxplain_vs_lewis_panel.py} & \path{lewis_audit_faithful/_individual} \\
Panel rule comparison (48 of 50; 24/7.3/15 vs 23/7.4/14) & \path{check_lewis_panel_definitions.py} & \path{lewis_audit_faithful/_individual} \\
\bottomrule
\end{longtable}
\endgroup

\subsection{Cross-class consensus}
\label{app:crossclass}

Table~\ref{tab:crossclass} reports frequency-based stability selection applied
to each model population separately and to all three pooled, at three stability
thresholds, under LIME. Each population contributes 10{,}000 individual fits
(100 splits $\times$ 100 fits), and each fit contributes its own top-50 genes.
Coverage is reported against the gene sets characterising the LIME partition of
the regularisation-path population: 16 genes shared between regimes, 16 specific
to the low-$C$ regime and 2 specific to the high-$C$ regime, each defined as
appearing in the regime's profile in at least 90\% of splits.

\begin{table}[htbp]
\centering
\small
\caption{Stability selection across model classes. ``Panel'' is the number of
genes retained at the stated threshold. Coverage columns give how many of the
shared, low-$C$-specific and high-$C$-specific gene sets the panel contains. The
pooled panel is never larger, and never recovers more of either regime's genes,
than the best single population at the same threshold.}
\label{tab:crossclass}
\begin{tabular}{@{}llcccc@{}}
\toprule
Threshold & Population & Panel & Shared (16) & Low-$C$-specific (16) & High-$C$-specific (2) \\
\midrule
\multirow{4}{*}{50\%}
 & path only    & 34 & 16 & 11 & 2 \\
 & network only & 27 & 13 & 6  & 1 \\
 & boosted only & 40 & 15 & 13 & 0 \\
 & all pooled   & 31 & 16 & 11 & 1 \\
\midrule
\multirow{4}{*}{70\%}
 & path only    & 23 & 16 & 5  & 2 \\
 & network only & 20 & 11 & 3  & 1 \\
 & boosted only & 31 & 14 & 13 & 0 \\
 & all pooled   & 19 & 13 & 6  & 0 \\
\midrule
\multirow{4}{*}{90\%}
 & path only    & 7  & 7  & 0  & 0 \\
 & network only & 10 & 9  & 1  & 0 \\
 & boosted only & 22 & 10 & 11 & 0 \\
 & all pooled   & 8  & 8  & 0  & 0 \\
\bottomrule
\end{tabular}
\end{table}
\subsection{Solver convergence}
\label{app:solver}
\begin{table}[htbp]
\centering
\caption{Convergence of the $10^4$ L2-regularised fits (lbfgs, iteration limit
1{,}000), by range of regularisation strength $C$, from refitting every fit in
the current environment and comparing it with the archived fit. No fit reached
the limit or raised a convergence warning; every refit reproduced the archived
held-out accuracy exactly, and its predicted probabilities to within
$5 \times 10^{-7}$. Each range holds 1{,}600 or 1{,}700 fits.}
\label{tab:solver_stability}
\begin{tabular}{lcc}
\toprule
$C$ range & max iterations & fits at limit \\
\midrule
$[10^{-3}, 10^{-2})$ & 50 & 0 \\
$[10^{-2}, 10^{-1})$ & 69 & 0 \\
$[10^{-1}, 1)$       & 70 & 0 \\
$[1, 10)$            & 47 & 0 \\
$[10, 10^{2})$       & 33 & 0 \\
$[10^{2}, 10^{3}]$   & 31 & 0 \\
\bottomrule
\end{tabular}
\end{table}
\end{document}